\documentclass[10pt,twocolumn,letterpaper]{article}

\usepackage[pagenumbers]{cvpr}

\usepackage{multirow}
\usepackage{xr}

\usepackage{cuted} 

\usepackage{currfile} 

\usepackage{caption} 
\definecolor{cvprblue}{rgb}{0.21,0.49,0.74}
\usepackage[pagebackref,breaklinks,colorlinks,allcolors=cvprblue]{hyperref}

\def\confName{CVPR}
\def\confYear{2027}

\title{Ref2Light: From One Image to Interactive Glasses-Free 3D Characters}

\author{
\begin{tabular*}{\textwidth}{@{\extracolsep{\fill}}ccc}
\begin{minipage}[t]{0.31\textwidth}
\centering
Hanzhong Zhang\\
Department of Computer Science\\
University of Exeter\\
Exeter, UK\\
{\tt\small armhiabelliard@gmail.com}
\end{minipage}
&
\begin{minipage}[t]{0.31\textwidth}
\centering
Haoyang Wang\\
Department of Mechanical Engineering\\
University of Alberta\\
Edmonton, AB, Canada\\
{\tt\small Haoyan25@ualberta.ca}
\end{minipage}
&
\begin{minipage}[t]{0.31\textwidth}
\centering
Siyang Song\thanks{Corresponding author: \texttt{s.song@exeter.ac.uk}.}\\
Department of Computer Science\\
University of Exeter\\
Exeter, UK\\
{\tt\small s.song@exeter.ac.uk}
\end{minipage}
\end{tabular*}
}


\begin{document}
\maketitle
\begin{strip}
\centering
\includegraphics[width=\linewidth]{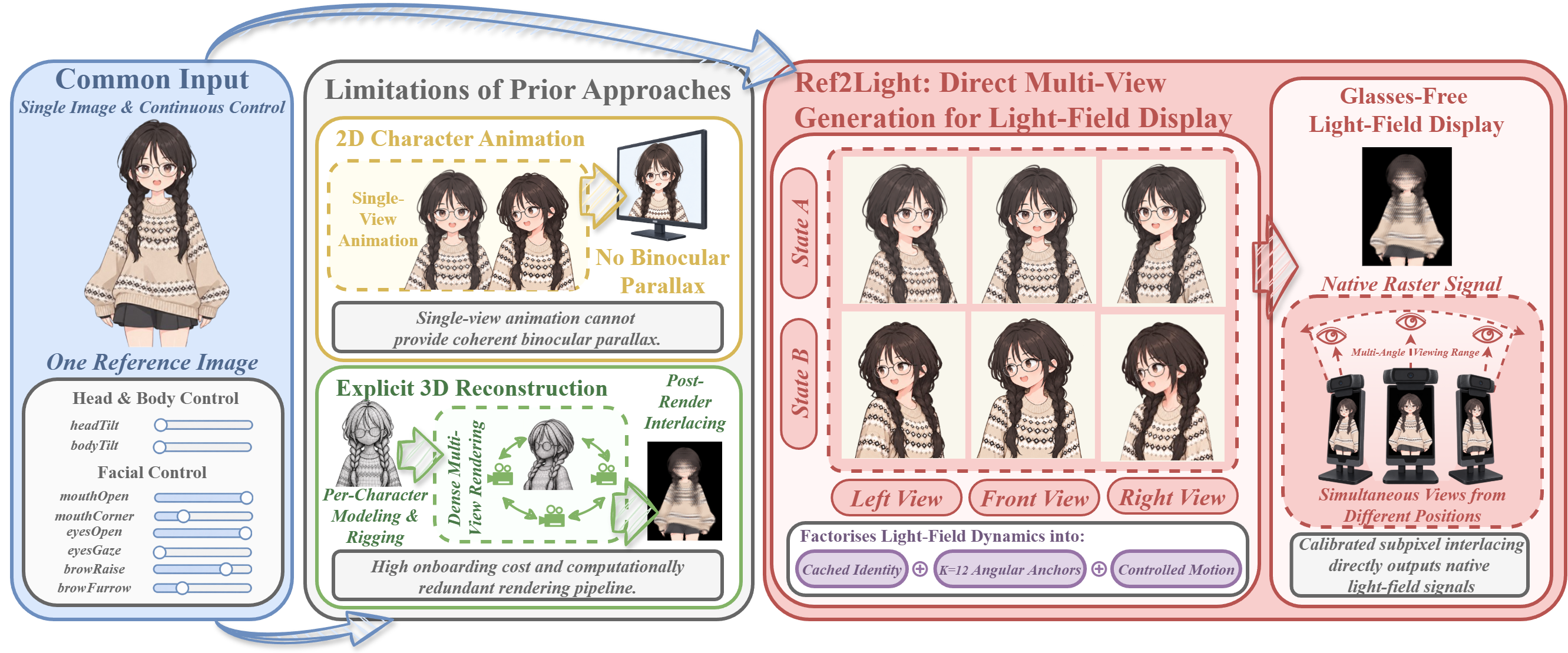}
\captionsetup{
    hypcap=false,
    skip=-1pt
}
\captionof{figure}{
Conceptual overview of Ref2Light. Given a single 2D illustration and continuous pose and expression controls, 2D character animators produce expressive motion from a single viewing direction, with no binocular parallax. Explicit 3D pipelines provide multiple views, but at the cost of per-character modeling and redundant dense rendering. Light-field displays offer a compelling alternative for presenting characters: by emitting different views toward different directions, they deliver glasses-free 3D with binocular and motion parallax for multiple viewers simultaneously, yet demand dense, consistent multi-view content. Ref2Light directly generates these views from a single reference image by factorizing light-field dynamics into cached identity, angular anchors, and controlled motion. Calibrated subpixel interlacing then converts the views into native display signals, enabling real-time glasses-free 3D character animation.
}
\label{fig:teaser}
\captionsetup{hypcap=true}
\end{strip}

\begin{abstract}
Animating character illustrations on light-field displays demands coherent binocular parallax, real-time pose articulation, and fine-grained expression control. However, existing 2D animation and 3D reconstruction pipelines cannot combine these with rapid character switching. We present Ref2Light, a feed-forward framework that maps a single A-pose illustration and continuous pose and expression controls to display-native light-field animation, without per-character 3D reconstruction, rigging, or test-time optimization. Our key insight is to decompose light-field synthesis into static identity, view-dependent parallax, and control-driven articulation. A shared encoder caches multi-scale identity features once per character for rapid switching. A conditional decoder then synthesizes ordered angular anchors from these features via backward texture transport and direct color synthesis. A global-local control scheme deforms the anchors with head-body controls and uses the updated correspondence fields to track facial regions across views for masked local facial refinement. Finally, calibrated differentiable subpixel interlacing converts the anchors into the native panel raster, enabling end-to-end native supervision without rendering all views at full resolution. On held-out VRoid identities, Ref2Light reduces head-body LPIPS by 20.1\%–55.6\% over 2D and 3D baselines and lowers non-target change MAE in mouth closure by 95.6\%–96.5\% versus monolithic decoding, while running at 34.0–40.1 FPS for 1440×2560 native rasters on a consumer GPU. Code and demo are in the Supplementary Material.
\end{abstract}

\section{Introduction}
\label{sec:intro}

\providecommand{\methodname}{Ref2Light}

Autostereoscopic light-field displays project glasses-free 3D parallax from a flat panel by routing the given RGB subpixels towards different viewing directions through a lenticular array~\cite{yang2024directl,wang2024naked}.
These displays provide a compelling medium for interactive digital characters in applications ranging from medical education~\cite{nguyen2025interactive} and telepresence avatars~\cite{fang2026lentiavatar} to virtual companions and conversational characters~\cite{arroyo2019bringing,wampfler2025platform}. Since slow preparation and delayed character responses can disrupt live human user–character interaction, these applications call for a workflow that bypasses costly 3D asset production by starting directly from a single 2D character illustration, i.e., the required workflow is expected to support switching between characters with minimal computational overhead, alongside continuous, fine-grained control of head pose, upper-body motion, and facial expressions for real-time interaction.

Although light-field content generation and rendering have been widely developed~\cite{song2025eye,sim2026coherentraster}, they generally focus on static view synthesis or efficient rendering of existing 3D scenes, leaving continuous digital character motion animation from a single 2D illustration unaddressed.
To achieve controllable character animation under continuous, dynamic control signals, three parallel technical routes have been frequently explored, with each offering distinct advantages in how it models the reference appearance, articulated motion and viewpoint dependence: 
(i) \textbf{monocular single-image 2D animators} transfer target head pose and facial expression controls directly to a 2D reference appearance through spatial warping~\cite{khungurn2025talking}, motion retargeting~\cite{guo2024liveportrait}, or disentangled editing~\cite{pang2023dpe}, which allows instant single-portrait onboarding and expressive 2D motion control, with lightweight architectures further enabling real-time inference efficiency~\cite{jiang2025mobileportrait}; 
(ii) \textbf{subject-specific 3D avatar} approaches reconstruct explicit or implicit 3D representations, including neural implicit fields~\cite{bai2024efficient} and 3D Gaussian avatars~\cite{qian2024gaussianavatars}, from multi-view captures or video to provide geometrically consistent viewpoint-dependent rendering. Within this route, light-field avatar systems further employ subject-specific driver distillation to achieve real-time multi-view presentation across the display viewing cone~\cite{fang2026lentiavatar}; and 
(iii) \textbf{feed-forward single-image 3D generators} exploit multi-view diffusion priors or large reconstruction models to infer 3D meshes or Gaussian assets directly from a single reference image~\cite{peng2024charactergen,qiu2025lhm}.
Compared with subject-specific 3D reconstruction, this route eliminates multi-view capture while preserving explicit 3D viewpoint consistency.

Despite such advances, simultaneously achieving rapid character onboarding, fine-grained facial expression control, and efficient multi-view synthesis for continuously driven interactive light-field characters remains an open challenge, as none of the above three routes satisfies all three requirements at once. Continuously driven interactive light-field characters remain an open challenge, as none of the above three routes satisfies all three requirements at once.
Specifically, although monocular single-image 2D animators offer fast onboarding and direct image-space control, they generate only a single-view output. Directly broadcasting this single view across display viewpoints eliminates binocular parallax, resulting in a flat appearance devoid of stereoscopic 3D depth (\textbf{Problem 1}).
While subject-specific 3D avatar pipelines resolve this cross-view inconsistency through shared 3D geometry, each new identity requires per-subject capture and 3D fitting, followed by subject-specific driver distillation for real-time display~\cite{fang2026lentiavatar}. 
Meanwhile, although feed-forward generators remove the multi-view capture step while preserving 3D consistency, they still need to infer a per-character 3D asset and construct its rigging or motion representation before driving~\cite{peng2024charactergen,qiu2025lhm,chu2024generalizable,zhang2025guava}. Consequently, both routes require a time-consuming and complex character-specific asset preparation process when onboarding a new character image (\textbf{Problem 2}).
Furthermore, even a character-specific asset is prepared, existing light-field display pipelines are required to render all full-resolution views prior to subpixel interlacing, incurring large computational and memory redundancy as only a sparse fraction of samples from each view contributes to the final 
panel raster (\textbf{Problem 3}).
In addition, despite that display-native renderers eliminate this full-view rendering overhead for pre-built 3D scenes~\cite{yang2024directl,sim2026coherentraster}, they require an existing 3D scene representation as the input (i.e., still suffering from the onboarding bottleneck of Problem~2).


Within a calibrated viewing cone, human observers typically receive neighboring views that share character identity and control state but differ in viewpoint~\cite{yang2024directl,wang2024naked}. Building on this, we decompose controllable light-field synthesis into static identity, view-dependent parallax, and control-driven articulation (illustrated in Fig.~\ref{fig:teaser}). Based on this decomposition, this paper proposes \methodname{}, a novel feed-forward framework that maps 
\textbf{a single frontal character illustration with continuous head-body and facial controls directly to a display-native light-field raster without per-character 3D reconstruction, rigging, or full-view rendering.}
Specifically, \methodname{} resolves the above three bottlenecks through a tightly coupled three-stage design. 
\textbf{First}, to eliminate the onboarding bottleneck (\textbf{addressing Problem~2}), a shared encoder extracts and caches multi-scale static identity features once from the reference image, providing a reusable appearance basis across all views and driving states. 
\textbf{Second}, to restore coherent binocular parallax (\textbf{addressing Problem~1}), a pose-conditioned decoder lifts these cached features into compact angular anchors across the viewing cone by coupling backward texture transport with direct disocclusion synthesis. 
\textbf{Third}, to achieve fine-grained expression control and bypass redundant dense-view rendering (\textbf{addressing Problem~3}), a correspondence-guided refinement module reuses the backward transport fields from the global anchors to track and edit local facial regions in-place, from which calibrated differentiable subpixel interlacing directly samples the native panel raster for end-to-end supervision.
Our main contributions are summarised as follows:
\begin{itemize}
    \item \textbf{A compact, controllable angular-anchor light-field representation.} We factorise dynamic multi-view character synthesis into cached static identity features and pose-conditioned hybrid angular anchors. \textbf{This differs from} monocular 2D animators that lack stereoscopic parallax and explicit 3D pipelines that require costly per-character reconstruction and rigging, enabling instant single-image onboarding alongside view-consistent posture control.

    \item \textbf{Correspondence-guided global--local control decomposition.} We introduce an in-place facial refinement mechanism that reuses global backward correspondence fields to dynamically track and edit facial regions within the angular anchors, coupled with pose-conditioned mouth-closure correction. \textbf{This differs from} monolithic control decoding that entangles facial expressions with global head pose, significantly improving closure accuracy while eliminating non-target texture drift.

    \item \textbf{Differentiable display-native synthesis and supervision.} We formulate optical subpixel interlacing as direct trilinear native-raster sampling over the refined angular anchors, closing the end-to-end training loop with native-panel supervision. \textbf{This differs from} conventional light-field pipelines that render all full-resolution views prior to interlacing or rely on pre-built 3D scenes, substantially reducing inference latency and memory footprint.
\end{itemize}

\section{Related Work}
\label{sec:related}

\textbf{Light-Field Displays:} 
Lenticular light-field displays route subpixels towards distinct viewing zones for glasses-free 3D~\cite{yang2024directl}. Recent work improves display quality via calibrated lens compensation~\cite{zhou2021compensation} and learned view-aperture encoding~\cite{chao2022time}. Efficient rendering pipelines use tailored projections~\cite{fink2023efficient}, display-aware point rendering~\cite{gavane2025light}, direct radiance rendering~\cite{yang2024directl}, and subpixel Gaussian rasterisation~\cite{sim2026coherentraster} to reduce dense-view rendering overhead. These methods focus on rendering existing scene representations.
For content generation, layered representations lift single images into small-baseline light fields~\cite{li2020synthesizing}, while EYE3 converts multimodal inputs via generative completion~\cite{song2025eye}. Both focus on view synthesis without semantic pose or expression control. For dynamic avatars, LentiAvatar reconstructs Gaussian heads from monocular video and accelerates driving through subject-specific distillation~\cite{fang2026lentiavatar}. Complementary work uses mesh-anchored Gaussians and spatial-angular attention for dense-view consistency~\cite{cui2026multi}. Our method couples single-reference character animation with differentiable native-panel interlacing, without per-character reconstruction.

\textbf{Single-Image Multi-View and 3D Generation:} 
Narrow-cone view synthesis from a single image uses representations including Multiplane Images~\cite{tucker2020single}, Layered Depth Images~\cite{shih20203d}, and variable-depth occlusion networks~\cite{li2020synthesizing}. These approaches primarily address viewpoint changes without semantic character articulation. Single-image 3D generation uses multi-view diffusion priors~\cite{liu2023zero,shi2024mvdream} and large reconstruction models~\cite{hong2024lrm,tang2024lgm}. Character-centric pipelines canonicalise illustrations into A-pose meshes~\cite{peng2024charactergen}, decompose semantic layers~\cite{he2025stdgen,chen2023panic}, or reconstruct animatable 3D Gaussians for humans~\cite{qiu2025lhm}. These pipelines introduce intermediate 3D assets before animation and multi-view rendering. In contrast, \methodname{} synthesizes controllable angular views directly from image features, avoiding per-character 3D reconstruction.

\textbf{Single-Image Character Animation:} 
Monocular animators decouple appearance from motion through keypoint-based warping~\cite{zhao2022thin}, 3D canonicalisation~\cite{wang2021one,drobyshev2022megaportraits}, retargeting~\cite{guo2024liveportrait}, cached inference~\cite{jiang2025mobileportrait}, and disentangled control~\cite{pang2023dpe,khungurn2025talking}. Pose-guided diffusion also enables character video generation~\cite{hu2024animate}. These methods do not directly target calibrated display-native outputs. Complementary 3D-aware approaches use triplane-based radiance fields~\cite{li2023one,trevithick2023real} or Gaussian avatars~\cite{chu2024generalizable,zhang2025guava}; template-based pipelines draw on parametric models such as FLAME~\cite{li2017learning} and SMPL-X~\cite{pavlakos2019expressive}. Our approach decouples identity encoding from driving and performs coordinate transport and local synthesis in angular-anchor space, directly sampling the native raster without rendering every display view.

\begin{figure*}[t]
    \centering
    \includegraphics[width=\textwidth]{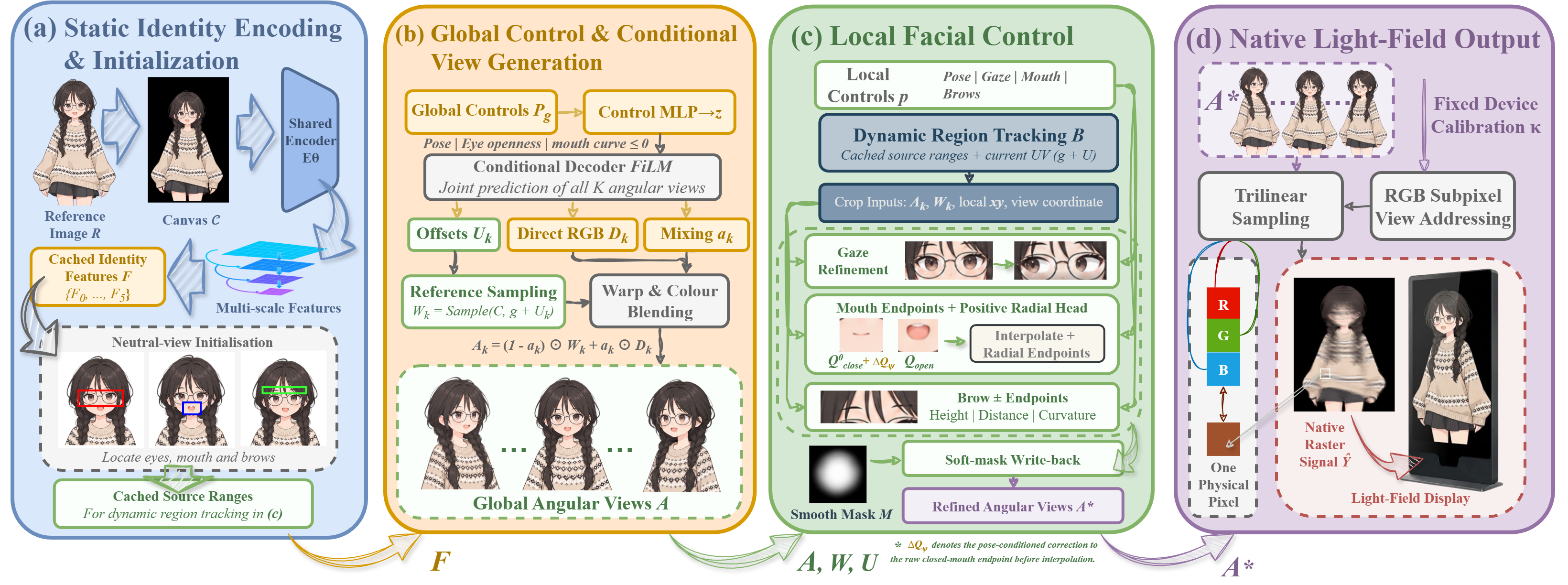}
    \captionsetup{
        skip=-1pt
    }
    \caption{
    \textbf{Overview of the proposed Ref2Light.}
    \textbf{(a) Static Identity Encoding \& Initialization:} Reference image $R$ is centred on canvas $C$ and encoded into cached multi-scale features $F$, while neutral views initialise facial boxes and cached source coordinate ranges on $C$.
    \textbf{(b) Global Control \& Conditional View Generation:} Global controls $p_g$ modulate a conditional decoder via FiLM to predict backward offsets $U_k$, direct RGB colours $D_k$, and mixing weights $\alpha_k$, blending warped reference textures $W_k$ with $D_k$ into global angular views $A$.
    \textbf{(c) Local Facial Control:} Dynamic region tracking $\mathcal{B}$ matches updated correspondences $g+U_k$ against cached source ranges to extract facial crops, where local controllers and pose-conditioned closure correction $\Delta Q_p$ refine gaze, mouth, and brows via soft-mask write-back into $A^*$.
    \textbf{(d) Native Light-Field Output:} Fixed calibration $\kappa$ guides RGB subpixel view addressing and trilinear sampling from $A^*$ directly into the native raster signal $\widehat{Y}$.}
    \vspace{-10pt}
    \label{fig:framework}
\end{figure*}

\section{Task Formulation}
\label{sec:task_definition}

Since light-field displays consume a single subpixel-interlaced signal for glasses-free 3D presentation, the task of \emph{display-native light-field character animation} is to learn a feed-forward model $\mathcal{F}$ that synthesizes a display-native RGB raster $\widehat{Y}$ preserving character identity, multi-view parallax, and pose-expression articulation from a single frontal A-pose RGB image $R$ (upright, arms angled downwards) and a per-frame control vector $p\in\mathbb{R}^{14}$ as:
\begin{equation}
    \widehat{Y}=\mathcal{F}(R,p;\kappa)\in[0,1]^{3\times H_n\times W_n},
\label{eq:task_definition}
\end{equation}
where $H_n,W_n$ denote the native-raster dimensions and $\kappa$ specifies fixed optical panel calibration.
Using zero-based indexing ($p_{i:j}$ denoting entries $i,\ldots,j-1$), $p$ parameterises 3D head and body rotations ($p_{0:3},p_{3:6}$, in degrees), eye and mouth openness ($p_6,p_7$), signed mouth curvature ($p_8$), brow height, spacing, and curvature ($p_{9:12}$), and 2D gaze ($p_{12:14}$).

\textbf{Challenge:} Directly learning $\mathcal{F}$ as a monolithic native-space network is ill-posed and computationally prohibitive. This is because spatially adjacent native subpixels address disparate viewing directions, unseen oblique geometry must be hallucinated consistently across views, and rendering all $V$ display views at full resolution incurs severe redundancy.

\section{Ref2Light: Display-Native Light-Field Character Animation}
\label{sec:method}

\textbf{Method overview.} To preserve view-space locality while bypassing full $V$-view rendering, our Ref2Light factorises $\mathcal{F}$ over $K$ ordered angular anchors 
($K<V$, with $K=12$ and $V=40$ across a ${\sim}40^\circ$ viewing cone, Sec.~\ref{sec:supp_anchor_budget}).
After a deterministic operator $\mathcal{P}$ centres $R$ onto a canonical canvas $C=\mathcal{P}(R)\in[0,1]^{3\times H_r\times W_r}$ (Sec.~\ref{sec:supp_input}), we decompose control $p$ into global posture $p_g$ and local facial expressions $p_l$, formulating Eq.~\eqref{eq:task_definition} as a four-stage composition matching Fig.~\ref{fig:framework}(a)--(d):
\begin{equation}
\begin{aligned}
    F &= E_\theta(C),\qquad
    (A,W,U) = D_\theta(F,p_g),\\
    A^* &= \mathcal{L}_{\phi,\psi}(A,W,U,p_l;\mathcal{B}),\qquad
    \widehat{Y} = \mathcal{S}_\kappa(A^*).
\end{aligned}
\label{eq:overview_comp}
\end{equation}
Specifically:
(1) \emph{Shared Encoding and Initialisation} ($E_\theta$, Sec.~\ref{sec:identity}) decouples static character appearance from motion by caching multi-scale identity features $F$ and canonical facial source ranges $\Omega_r$ once at onboarding;
(2) \emph{Conditional View Generation} ($D_\theta$, Sec.~\ref{sec:global_gen}) synthesises pose-conditioned angular views $A$ via hybrid backward texture transport ($W,U$) and direct disocclusion synthesis;
(3) \emph{Local Facial Control} ($\mathcal{L}_{\phi,\psi}$, Sec.~\ref{sec:local_ctrl}) reuses the backward correspondence field $U$ to dynamically track facial boxes $\mathcal{B}$ across views, confining fine-grained expression edits and pose-aware mouth closure to $A^*\in[0,1]^{K\times 3\times H_a\times W_a}$; and
(4) \emph{Native Light-Field Output} ($\mathcal{S}_\kappa$, Sec.~\ref{sec:native_output}) performs calibrated subpixel trilinear sampling from $A^*$ into $\widehat{Y}$.

\subsection{Static Identity Encoding \& Initialization}\label{sec:identity}

Instant character onboarding requires isolating everything the model knows about a character's static identity from time-varying driving signals and viewing directions.
As shown in Fig.~\ref{fig:framework}(a), a shared encoder $E_\theta$ maps $C$ into cached multi-scale feature maps $F=\{F_s\}_{s=0}^{5}$ that are strictly invariant to control $p$ and anchor index $k\in\{1,\ldots,K\}$:
\begin{equation}
    F=E_\theta(C),\qquad
    \frac{\partial F}{\partial p}=\frac{\partial F}{\partial k}=0.
\label{eq:identity}
\end{equation}
During continuous driving, downstream modules read this static appearance basis via U-Net skip connections~\cite{ronneberger2015unet}, amortising encoding overhead across frames.

\textbf{Canonical correspondence anchoring.}
To track facial regions across views and head poses during driving without invoking an object detector every frame, we anchor each region $r\in\{\text{left/right eye, brows, mouth}\}$ in the canonical coordinate space of $C$ during onboarding.
Evaluating $D_\theta$ once under neutral controls $p_{\mathrm{neu}}$ produces neutral views $\bar A_k$ and backward offsets $\bar U_k$.
Detecting bounding boxes $\bar b_r^k$ on $\bar A_k$ and projecting their normalised target coordinates $g(x)\in[-1,1]^2$ onto $C$ via $\bar U_k$ defines the cached source range:
\begin{equation}
    \Omega_r=\big\{\,g(x)+\bar U_k(x)\;\big|\;x\in \bar b_r^k\,\big\}\subset[-1,1]^2,
\label{eq:init_range}
\end{equation}
together with reference box dimensions $s_r$ (Sec.~\ref{sec:supp_tracking}).

\subsection{Global Control \& Conditional View Generation}\label{sec:global_gen}

Synthesising multi-view parallax under global posture changes faces a fundamental trade-off: pure spatial warping preserves sharp reference details but fails in disoccluded regions, whereas pure direct regression blurs high-frequency identity textures.
We resolve this by formulating conditional view generation as a hybrid transport-and-synthesis process across the $K$ angular anchors (Fig.~\ref{fig:framework}(b)).

Specifically, the global control vector $p_g\in\mathbb{R}^{14}$ retains head/body rotations ($p_{0:6}$), eye openness ($p_6$), and non-positive mouth curvature ($\min(p_8,0)$), while neutralising local expression axes.
Injected into decoder $D_\theta$ via a control MLP and bounded FiLM modulation~\cite{perez2018film} (Sec.~\ref{sec:supp_global}), $p_g$ conditions $F$ to jointly predict 2D backward correspondence offsets $U_k$, direct disocclusion RGB maps $D_k$, and pixel-wise blending weights $\alpha_k$ for all $K$ anchors:
\begin{equation}
\begin{aligned}
    \big\{(U_k,\,D_k,\,\alpha_k)\big\}_{k=1}^{K} &= D_\theta(F,\,p_g),\\
    A_k = (1-\alpha_k)\odot W_k &+ \alpha_k\odot D_k,
\end{aligned}
\label{eq:decoder_blend}
\end{equation}
where $W_k=\operatorname{Sample}(C,\operatorname{clip}(g+U_k,-1.2,1.2))$ transports reference appearance from $C$ via differentiable bilinear sampling~\cite{jaderberg2015spatial} and $\odot$ denotes element-wise multiplication.
By construction, $W_k$ anchors cross-view identity consistency by copying visible surfaces from $C$, while $D_k$ hallucinates newly exposed geometry at oblique angles.
Furthermore, $U_k$ establishes dense view- and pose-dependent correspondences back to $C$, and $(A,W,U)$ can be cached whenever $p_g$ is stationary.

\subsection{Local Facial Control}\label{sec:local_ctrl}

Routing all 14 controls through a single global decoder entangles subtle facial expressions with global posture, causing non-target texture drift and inaccurate mouth closure under head rotation.
Instead, our local operator $\mathcal{L}_{\phi,\psi}$ exploits the backward correspondences $U_k$ to track and refine facial regions in-place within the posed anchors $A$ (Fig.~\ref{fig:framework}(c)).

\textbf{Correspondence-driven region tracking.}
Because $g(x)+U_k(x)$ maps any pixel $x$ in anchor $k$ to its canonical source coordinate in $C$, matching this field against the cached source range $\Omega_r$ immediately localises region $r$ under arbitrary head-body poses and viewing angles:
\begin{equation}
    S_r^k=\big\{x\,\big|\,g(x)+U_k(x)\in\Omega_r\big\},\quad
    b_r^k=\operatorname{Box}\!\big(\operatorname{ctr}(S_r^k),s_r\big),
\label{eq:track}
\end{equation}
where $\operatorname{ctr}(S_r^k)$ extracts the tracked region centroid to place bounding box $b_r^k\in\mathcal{B}$ (translating eye/mouth boxes and adapting brow scale; Sec.~\ref{sec:supp_tracking}).

\textbf{Masked regional articulation.}
From each tracked box $b_r^k$, we crop current anchor appearance $B_{\mathrm{crop}}$ from $A_k$ and transported reference texture $W_{\mathrm{crop}}$ from $W_k$.
Conditioned on these crops, normalised local coordinates $g_{\mathrm{loc}}$, viewing angle, and local controls $p_l$, shared regional controllers $\phi$ predict a local 2D flow $u$ and RGB residual $r$ to compute a spatially gated update $\Delta$:
\begin{equation}
    \Delta=M\odot a\,\big[\operatorname{Sample}(B_{\mathrm{crop}},\,g_{\mathrm{loc}}+u)+r-B_{\mathrm{crop}}\big],
\label{eq:local}
\end{equation}
where $a$ is a scalar activation gate (e.g., clipped $L_1$ gaze magnitude scaled by eye openness) and $M\in[0,1]$ is a smooth spatial mask vanishing at crop borders.
Writing the resized update $\Delta$ back into $A_k$ guarantees zero modification outside $\mathcal{B}$.

\textbf{Pose-conditioned mouth closure.}
Because the 2D projection of a closing mouth varies significantly with 3D head orientation, interpolating between pose-agnostic open and closed templates produces misalignment under pitched or yawed poses.
We therefore introduce a zero-initialised correction head $H_\psi$ conditioned on the six head-body angles $p_{0:6}$, local decoder features $F_{\mathrm{mouth}}$, and crops $B_{\mathrm{mouth}},W_{\mathrm{mouth}}$ to adapt the raw 5-channel closed endpoint $Q_{\mathrm{close},0}$ (2D flow plus RGB residual) before linear interpolation with the open endpoint $Q_{\mathrm{open},0}$:
\begin{equation}
\begin{aligned}
    Q_{\mathrm{mouth}} &= (1-t)\,(Q_{\mathrm{close},0}+\Delta Q_p)+t\,Q_{\mathrm{open},0},\\
    \Delta Q_p &= H_\psi(F_{\mathrm{mouth}},\,B_{\mathrm{mouth}},\,W_{\mathrm{mouth}},\,p_{0:6}),
\end{aligned}
\label{eq:mouth_pose}
\end{equation}
where $t=\operatorname{clip}(p_7/u_{\max},0,1)$ normalises mouth opening $p_7$.
Mapping $Q_{\mathrm{mouth}}$ via a scaled $\tanh$ provides $(u,r)$ in Eq.~\eqref{eq:local}; as $t\to 1$, $(1-t)\Delta Q_p\to 0$, concentrating pose correction strictly on closure states.
Applying positive radial expansion ($p_8>0$) and signed brow refinement yields the final articulated anchor volume $A^*$ (Sec.~\ref{sec:supp_local}).

\subsection{Native Light-Field Output}\label{sec:native_output}

To bridge the $K$ articulated anchors $A^*\in[0,1]^{K\times 3\times H_a\times W_a}$ and the physical panel without materialising all $V$ full-resolution display views, the calibrated sampler $\mathcal{S}_\kappa$ formulates optical subpixel interlacing as a direct, parameter-free trilinear lookup (Fig.~\ref{fig:framework}(d)).
Under the fixed lenticular calibration $\kappa$, each physical pixel $(x,y)$ and colour subpixel $c\in\{0,1,2\}$ is routed to a specific viewing direction, which we map to normalised spatial coordinates $g^x(x),g^y(y)\in[-1,1]$ and a continuous normalised angular coordinate $g_c^z(x,y)\in[-1,1]$ along the $K$-anchor axis (exact phase-to-view mapping in Sec.~\ref{sec:supp_interlace}).
The sampler then queries each native subpixel directly from the anchor volume in FP32 via $\widehat{Y}_c(x,y)=\operatorname{TriSample}(A^*_c,g^x(x),g^y(y),g_c^z(x,y))$.
Because $V>K$, subpixels addressing intermediate viewing zones smoothly interpolate across adjacent anchors, while full differentiability with respect to $A^*$ allows native-raster supervision to back-propagate end-to-end into both local and global synthesis modules.

\subsection{Training Supervision and Optimisation}\label{sec:supervision}

End-to-end training must supervise the ill-posed decomposition between texture transport ($U$) and disocclusion synthesis ($D$) while preventing the model from leaking the reference image's initial expression into driven states.

\textbf{Decomposed transport-synthesis supervision.}
Using rendered ground-truth (GT) views $I^{\mathrm{gt}}$, native rasters $Y$, and projected canonical backward offsets $U^{\mathrm{gt}}$, we construct disjoint binary masks $M_f$ for copyable surface regions and $M_d$ for disoccluded direct-colour regions ($M_d\neq 1-M_f$).
The paired supervision objective is:
\begin{equation}
    \mathcal{L}_{\mathrm{pair}}=\mathcal{L}_{\mathrm{rgb}}+\lambda_f\ell_{M_f}(U,U^{\mathrm{gt}})+\lambda_d\ell_{M_d}(D,I^{\mathrm{gt}})+\lambda_n\mathcal{L}_{\mathrm{native}},
\label{eq:loss_pair}
\end{equation}
where $\ell_M$ denotes mask-normalised $L_1$ error, and $\mathcal{L}_{\mathrm{rgb}},\mathcal{L}_{\mathrm{native}}$ measure view-space and foreground-weighted native-raster reconstruction errors against $I^{\mathrm{gt}}$ and $Y$.
We evaluate $\mathcal{L}_{\mathrm{pair}}$ over neutral ($\mathcal{L}_o$), geometrically augmented ($\mathcal{L}_a$), and expression-varied ($\mathcal{L}_e$) references (Sec.~\ref{sec:supp_pair}).

\textbf{Cross-reference response matching.}
To eliminate source-expression leakage when driving non-neutral references $R_a$, a replay loss $\mathcal{L}_{\mathrm{rep}}$ enforces consistency between the control-induced facial RGB differential $\Delta\mathcal{F}(R_a,p)=\mathcal{F}(R_a,p)-\mathcal{F}(R_a,p_{\mathrm{base}})$ and that of a frozen initial model $\mathcal{F}_{\mathrm{init}}$ evaluated on neutral reference $R_0$:
\begin{equation}
    \mathcal{L}_\Delta=\ell_{\mathrm{face}}\!\big(\Delta\mathcal{F}(R_a,p),\,\Delta\mathcal{F}_{\mathrm{init}}(R_0,p)\big),
\label{eq:loss_delta}
\end{equation}
which combines with native GT and identity-retention terms to form $\mathcal{L}_{\mathrm{rep}}$ (Sec.~\ref{sec:supp_training}).
We optimise the joint objective $\mathcal{L}_{\mathrm{global}}=0.2(\mathcal{L}_o+\mathcal{L}_a+\mathcal{L}_e)+0.4\mathcal{L}_{\mathrm{rep}}$ in three stages: pretraining and freezing $\phi$, warming up $\psi$ with $\theta$ fixed, and jointly training $(\theta,\psi)$ for 16k steps, followed by 2,000 steps of native-raster refinement ($\mathcal{L}_{\mathrm{native}}$) on mouth-open references with $E_\theta$ frozen.

\providecommand{\ours}{\textbf{Ref2Light}}
 
\section{Experiments}
\label{sec:experiments}
 
\subsection{Experimental Setup}
\label{subsec:exp_setup}
 
\noindent\textbf{Datasets and Pairing.}
All methods take a $512\times512$ reference and continuous controls, evaluated across 40 calibrated views ($540\times960$) and the interlaced $1440\times2560$ native raster ($V=40$, $K=12$ for \ours{}) with references and ground-truth (GT) views rendered from 3D assets. We use VRoid assets from the PAniC-3D collection and Genshin models obtained through its AnimeRecon download scripts~\cite{chen2023panic}. With fixed final weights (Sec.~\ref{sec:supp_training}), we evaluate two benchmarks with no identity overlap with Ref2Light's training set: (1) \textbf{VRoid} on 48 paired unseen identities across nine discrete states, with eight also evaluated on three 32-frame trajectories; (2) \textbf{zero-shot Genshin} on 22 paired identities (static/head-body), 24 (facial), and 10 (local closure). A separate nine-identity diagnostic tests 40 control states at three views.

\noindent\textbf{Baselines, Metrics, and Hardware.}
We compare \ours{} with monocular 2D baselines, \textbf{THA4 + Fixed Plane}~\cite{khungurn2025talking} and \textbf{THA4 + Monocular Depth} (pairing THA4 with monocular depth and calibrated reprojection; Sec.~\ref{sec:supp_depth}) across all task groups, and feed-forward 3D baselines, \textbf{LHM-1B}~\cite{qiu2025lhm} and \textbf{CharacterGen + Pose Adaptation}~\cite{peng2024charactergen}, on supported static and head-body axes. We report view-space FG-PSNR, SSIM\cite{wang2004ssim}, and LPIPS\cite{zhang2018lpips}; native-raster Nat-RMSE; active-mask closure MAE; GT-referenced Temporal and Angular Change MAE; and diagnostic target-region Change MAE and non-target Outside MAE, averaging metrics across states and views per identity before equal-weight identity aggregation (Sec.~\ref{sec:supp_metrics}). To evaluate real-world deployability on mobile workstations, runtime is measured on a single NVIDIA RTX 5090 Laptop GPU (24~GB) from CPU control dispatch to GPU-ready native RGB, where Sync FPS denotes the reciprocal of mean frame latency. 
 
\begin{table*}[t]
\centering
\caption{Quantitative comparisons on the in-domain VRoid and zero-shot cross-dataset Genshin benchmarks. Complete six-metric results are reported in Tables S1 and S7. Best results are in bold.}
\label{tab:main_quality}
\scriptsize
\setlength{\tabcolsep}{4.5pt}
\renewcommand{\arraystretch}{1.08}
\begin{tabular}{ll cccc c cccc}
\toprule
& & \multicolumn{4}{c}{\textbf{In-Domain Benchmark (VRoid, $N=48$ Paired Identities)}} & & \multicolumn{4}{c}{\textbf{Zero-Shot Cross-Dataset (Genshin, $N=22 / 24$ Paired)}} \\
\cmidrule(lr){3-6} \cmidrule(lr){8-11}
\textbf{Task Group} & \textbf{Method} & \textbf{FG-PSNR}$\uparrow$ & \textbf{SSIM}$\uparrow$ & \textbf{LPIPS}$\downarrow$ & \textbf{Nat-RMSE}$\downarrow$ & & \textbf{FG-PSNR}$\uparrow$ & \textbf{SSIM}$\uparrow$ & \textbf{LPIPS}$\downarrow$ & \textbf{Nat-RMSE}$\downarrow$ \\
\midrule
\multirow{5}{*}{\shortstack[l]{\textbf{Static}\\\textbf{Reconst.}}} 
& CharacterGen & 8.678 & 0.6903 & 0.4049 & 0.2569 & & -- & -- & -- & -- \\
& LHM-1B & 11.411 & 0.6847 & 0.4319 & 0.2485 & & 8.932 & 0.5404 & 0.5285 & 0.2912 \\
& THA4 + Fixed Plane & 10.971 & 0.7746 & 0.2402 & 0.1929 & & 8.825 & 0.6498 & 0.3508 & 0.2534 \\
& THA4 + Depth & 10.570 & 0.7883 & 0.2210 & 0.1934 & & 8.964 & 0.6792 & \textbf{0.3081} & 0.2383 \\
& \ours{} & \textbf{16.915} & \textbf{0.8405} & \textbf{0.1896} & \textbf{0.1174} & & \textbf{11.178} & \textbf{0.6831} & 0.3464 & \textbf{0.2014} \\
\midrule
\multirow{5}{*}{\shortstack[l]{\textbf{Head-Body}\\\textbf{Motion}}} 
& CharacterGen & 8.612 & 0.6872 & 0.4115 & 0.2594 & & -- & -- & -- & -- \\
& LHM-1B & 11.401 & 0.6850 & 0.4322 & 0.2485 & & 8.923 & 0.5409 & 0.5285 & 0.2912 \\
& THA4 + Fixed Plane & 10.778 & 0.7667 & 0.2551 & 0.1986 & & 8.731 & 0.6433 & 0.3673 & 0.2576 \\
& THA4 + Depth & 10.228 & 0.7741 & 0.2401 & 0.2013 & & 8.827 & 0.6687 & \textbf{0.3275} & 0.2436 \\
& \ours{} & \textbf{16.826} & \textbf{0.8390} & \textbf{0.1918} & \textbf{0.1183} & & \textbf{11.133} & \textbf{0.6817} & 0.3489 & \textbf{0.2026} \\
\midrule
\multirow{3}{*}{\shortstack[l]{\textbf{Eye, Mouth}\\\textbf{\& Gaze}}} 
& THA4 + Fixed Plane & 10.993 & 0.7751 & 0.2404 & 0.1926 & & 8.966 & 0.6528 & 0.3482 & 0.2497 \\
& THA4 + Depth & 10.570 & 0.7885 & 0.2217 & 0.1934 & & 9.070 & 0.6810 & \textbf{0.3081} & 0.2360 \\
& \ours{} & \textbf{16.920} & \textbf{0.8408} & \textbf{0.1896} & \textbf{0.1173} & & \textbf{11.240} & \textbf{0.6844} & 0.3475 & \textbf{0.1993} \\
\bottomrule
\end{tabular}
\vspace{-2mm}
\end{table*}
 
\subsection{Quantitative Results}
\label{subsec:quality_comparison}
 
\begin{figure}[t]
    \centering
    \includegraphics[width=\columnwidth]{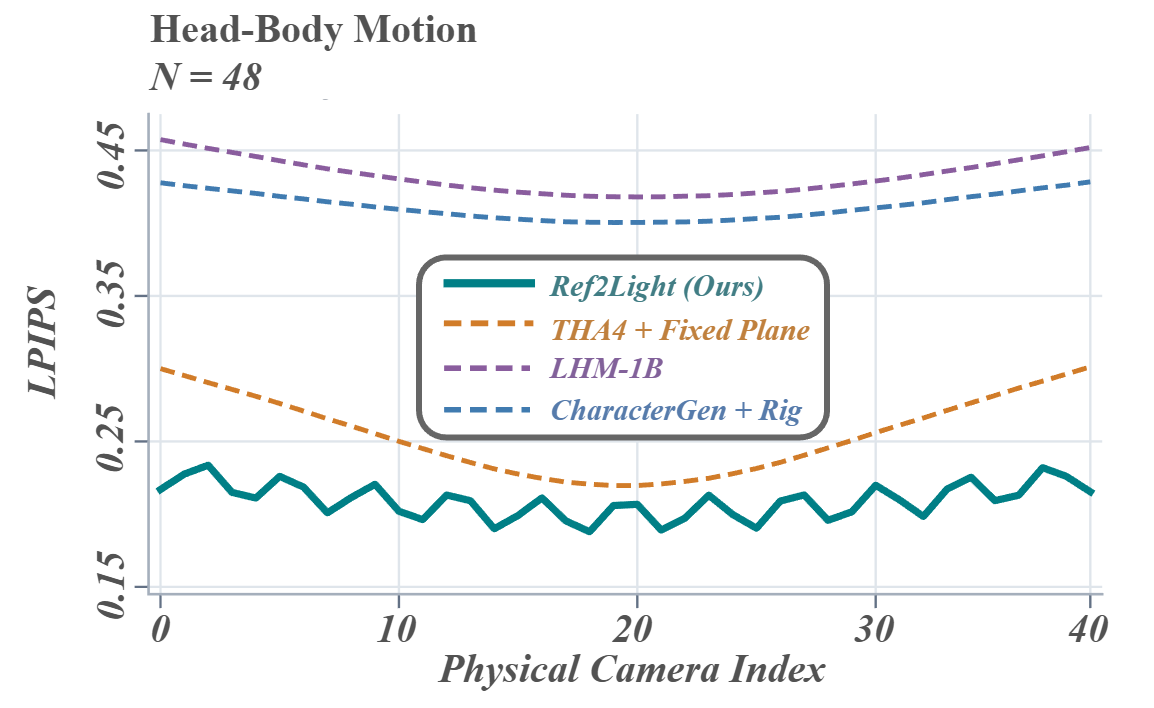}
    \vspace{-2mm}
    \caption{
        \textbf{Perceptual quality across viewing directions under head-body motion.}
        All four baselines, including THA4 + Depth, are included. LPIPS at each of the 40 physical camera positions is averaged over head-body control states within each identity and then with equal weight across the $N=48$ paired unseen VRoid identities. Lower values indicate better perceptual fidelity.
    }
    \label{fig:angular_and_qualitative}
    \vspace{-3mm}
\end{figure}

\noindent\textbf{In-Domain Performance.}
Table~\ref{tab:main_quality} shows that \ours{} leads all metrics across the three VRoid task groups. Under head-body motion, LPIPS and Nat-RMSE decrease by \textbf{20.12\%--55.62\%} and \textbf{40.42\%--54.39\%} over the four baselines. Figure~\ref{fig:angular_and_qualitative} confirms lower LPIPS across all 40 views.
 
\noindent\textbf{Zero-Shot Cross-Dataset Generalisation.}
On Genshin characters (Table~\ref{tab:main_quality}, right), \ours{} reduces head-body Nat-RMSE ($N=22$) by \textbf{16.84\%} ($19/22$ identities improved), \textbf{21.37\%}, and \textbf{30.44\%} over THA4 + Depth, THA4 + Fixed Plane, and LHM-1B, while lowering LPIPS by $5.01\%$ and $33.98\%$ over the latter two (despite THA4 + Depth retaining lower cross-domain LPIPS). On facial control ($N=24$), \ours{} also reduces Nat-RMSE by \textbf{15.55\%} over THA4 + Depth. Furthermore, on the 10-identity local evaluation set, Eye/Mouth Closure MAE reaches \textbf{0.1324/0.0812}, reducing fixed-plane THA4 errors by \textbf{31.34\%/67.91\%} ($10/10$ identities; Sec.~\ref{sec:supp_cross}).
 
\subsection{Qualitative Results}
\label{subsec:qualitative}
 
\begin{figure}[t]
    \centering
    \includegraphics[width=\columnwidth]{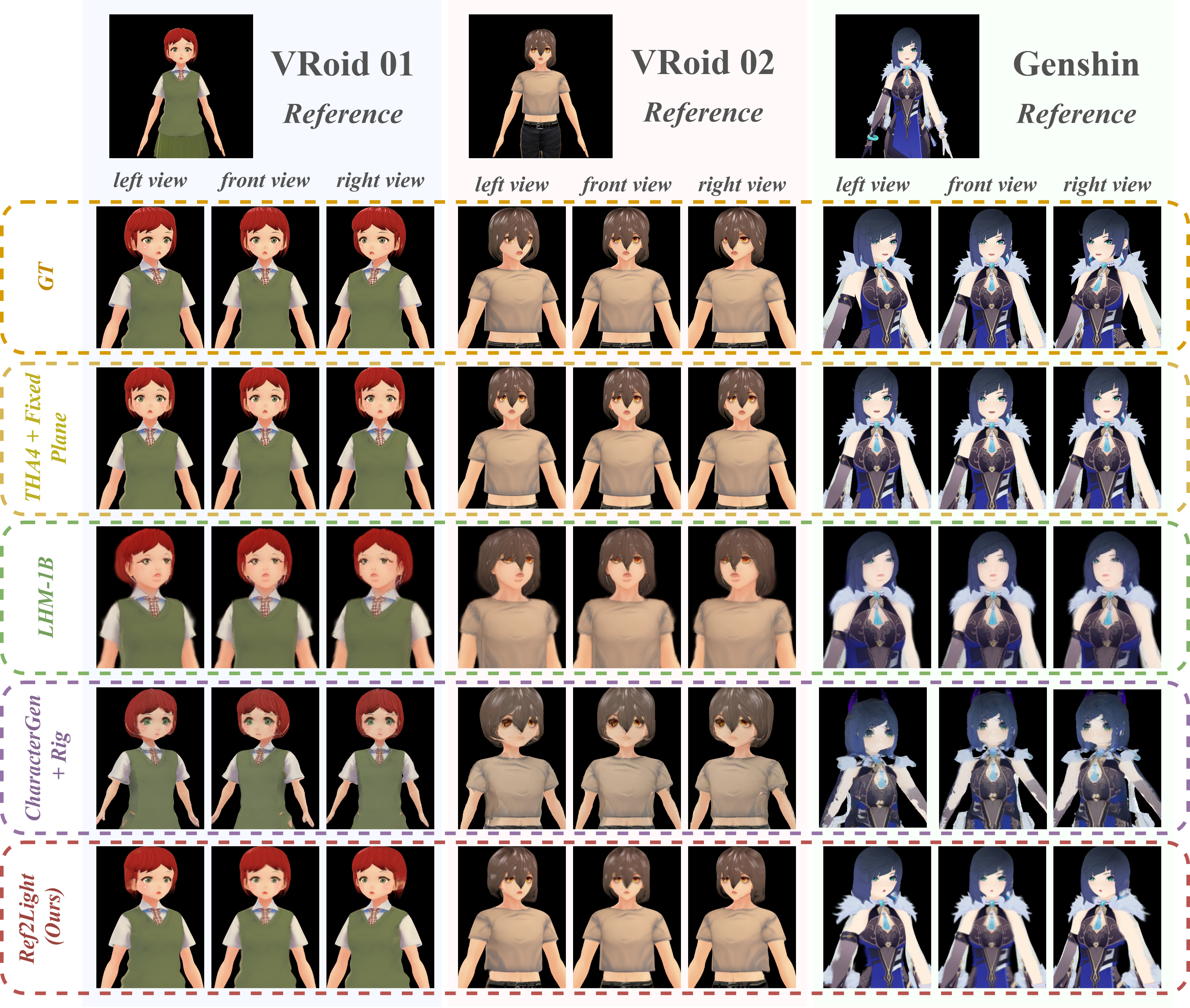}
    \caption{
        \textbf{Qualitative multi-view comparisons on VRoid and Genshin characters.}
        Each group presents the input reference illustration alongside GT and predicted views at left, frontal, and right viewing positions, with the character control state held fixed across viewpoints and methods. All rows use identical upper-body crops. The Genshin example shows zero-shot transfer; CharacterGen is not evaluated in that domain.
    }
    \label{fig:qualitative_main}
\end{figure}

\begin{figure*}[t]
    \centering
    \includegraphics[width=\textwidth]{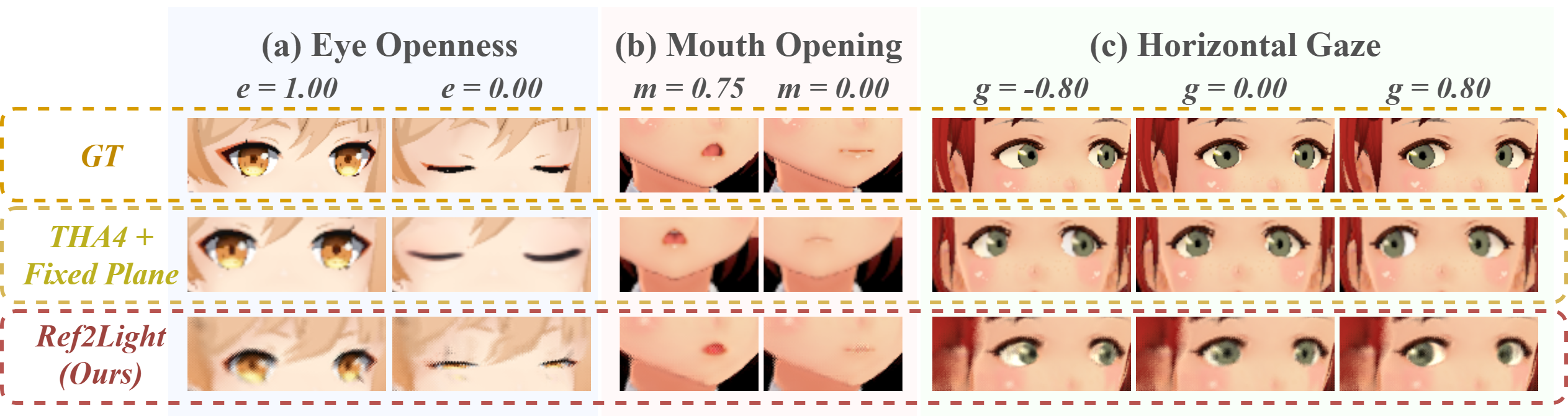}
    \caption{
        \textbf{Qualitative comparison of local facial control.}
        Close-up views of (a) eye openness $e$, (b) mouth opening $m$, and (c) horizontal gaze $g$. Rows show GT, THA4 + Fixed Plane, and Ref2Light; columns within each panel correspond to the indicated control values.
    }
    \label{fig:local_control}
\end{figure*}

Figure~\ref{fig:qualitative_main} compares GT and predicted left, frontal, and right views on unseen VRoid and Genshin characters. While THA4 + Fixed Plane lacks depth-dependent parallax and the illustrated 3D examples exhibit facial or body distortions, \ours{} preserves characteristic hairstyles, facial structure, and clothing across views. Fixed-scanline EPI diagnostics examine angular RGB structure against GT (Sec.~\ref{sec:supp_review_epi}).
Figure~\ref{fig:local_control} further illustrates eye-openness, mouth-opening, and horizontal-gaze responses alongside GT and THA4 + Fixed Plane.
 
\subsection{Continuous Driving and Efficiency}
\label{subsec:continuous_runtime}
\label{subsec:continuous_eval}
\label{subsec:runtime_efficiency}
 
\noindent\textbf{Continuous Driving.}
Across three 32-frame trajectories on eight unseen identities, \ours{} reduces head-body LPIPS by \textbf{22.51\%} over THA4 + Fixed Plane ($7/8$ identities), \textbf{55.05\%} over LHM-1B ($8/8$), and \textbf{52.58\%} over CharacterGen ($8/8$), while lowering facial and combined LPIPS by \textbf{19.75\%} and \textbf{22.61\%} over the fixed-plane baseline. Figure~\ref{fig:supp_continuous} visualises trajectory sequences, and Table~\ref{tab:continuous} reports full reconstruction and GT-referenced transition metrics.
 
\begin{table}[t]
\centering
\caption{Runtime and memory under different control trajectories.}
\label{tab:runtime}
\scriptsize
\setlength{\tabcolsep}{3.2pt}
\renewcommand{\arraystretch}{1.08}
\begin{tabular}{llccc}
\toprule
\textbf{Trajectory} & \textbf{Method} & \textbf{Sync ms}$\downarrow$ & \textbf{FPS}$\uparrow$ & \textbf{GiB}$\downarrow$ \\
\midrule
\multirow{2}{*}{\shortstack[l]{Eye, Mouth\\\& Gaze}} & THA4 + Fixed & 94.78 & 10.55 & 1.453 \\
 & \ours{} & \textbf{28.96} & \textbf{34.53} & \textbf{1.227} \\
\multirow{4}{*}{\shortstack[l]{Head-Body\\Motion}} & THA4 + Fixed & 94.65 & 10.57 & 1.453 \\
 & LHM-1B & 64.63 & 15.47 & 12.894 \\
 & CharacterGen & \textbf{22.81} & \textbf{43.84} & 2.724 \\
 & \ours{} & 24.94 & 40.09 & \textbf{1.226} \\
\multirow{2}{*}{\shortstack[l]{Combined\\Motion}} & THA4 + Fixed & 94.40 & 10.59 & 1.453 \\
 & \ours{} & \textbf{29.39} & \textbf{34.02} & \textbf{1.226} \\
\bottomrule
\end{tabular}
\end{table}
 
\noindent\textbf{Runtime, Precision, and Memory.}
Table~\ref{tab:runtime} indicates that \ours{} sustains \textbf{34.02--40.09 FPS} with \textbf{1.226~GiB} mean peak allocated memory during head-body driving, versus $12.894$~GiB for LHM-1B. In the head-body precision comparison (Table~\ref{tab:supp_review_precision}), BF16 THA4 + Fixed Plane reaches \textbf{13.60 FPS}, whereas \ours{} achieves \textbf{30.39 FPS} in eager mode (\textbf{2.23$\times$} faster) and \textbf{42.95 FPS} with optimised execution.
 
\noindent\textbf{Reference Switching and Native Sampling.}
With pre-loaded weights, switching among eight pre-aligned inputs takes \textbf{0.661~s} median. Given refined anchors, direct native sampling is \textbf{3.42$\times$} faster than expanding 40 full-resolution views before interlacing (Sec.~\ref{sec:supp_runtime}).
 
\begin{table}[t]
\centering
\caption{\textbf{Control decomposition and non-target preservation.} (a) Closure response and non-target preservation on the nine-identity diagnostic set. Change MAE measures control-induced RGB change error, while Outside MAE measures the error outside the target region. (b) Active-mask mouth-closure reconstruction on the formal benchmarks. Best results are in bold.}
\label{tab:ablation}
\scriptsize
\setlength{\tabcolsep}{3.2pt}
\renewcommand{\arraystretch}{1.08}
\begin{tabular}{@{}lcccc@{}}
\toprule
\multicolumn{5}{l}{\textbf{(a) Closure Response and
Non-Target Preservation ($N=9$)}} \\
\addlinespace[2pt]

& \multicolumn{2}{c}{Change MAE$\downarrow$}
& \multicolumn{2}{c}{Outside MAE$\downarrow$
($\times10^{-3}$)} \\

\cmidrule(lr){2-3}
\cmidrule(lr){4-5}

Control & Mono. & \ours{} & Mono. & \ours{} \\
\midrule

Frontal & 0.02946 & \textbf{0.02624}
& 4.701 & \textbf{0.163} \\

Posed & 0.03009 & \textbf{0.02730}
& 4.722 & \textbf{0.209} \\

\midrule

\multicolumn{5}{l}{\textbf{(b) Mouth-Closure Reconstruction}} \\
\addlinespace[2pt]

Domain & $N$ & Monolithic & \ours{} & Reduction$\uparrow$ \\
\midrule

VRoid & 48 & 0.04427 & \textbf{0.04281} & 3.30\% \\
Genshin & 24 & 0.11172 & \textbf{0.10691} & 4.31\% \\

\bottomrule
\end{tabular}
\end{table}
 
\subsection{Ablation Studies}
\label{subsec:ablation}
 
\noindent\textbf{Control Decomposition and Non-Target Preservation.}
We compare \ours{}  with a monolithic baseline under matched refinement. Both models start from the same 16k-stage parent checkpoint and receive 2,000 additional native-signal refinement updates on 418 training identities. The monolithic baseline routes all 14 controls through a single global decoder without correspondence tracking or masked local updates (Sec.~\ref{sec:supp_ablation}). We evaluate closure responsiveness and non-target preservation on an independent nine-identity diagnostic set, and mouth-closure reconstruction on the formal benchmarks. On the nine-identity diagnostic benchmark (fixing one reference-state translation per model and view across 40 states; Table~\ref{tab:ablation}(a)), \ours{} reduces frontal and posed closure Change MAE by \textbf{10.94\%} and \textbf{9.28\%} ($8/9$ identities improved) and Outside MAE by \textbf{96.52\%} and \textbf{95.56\%} ($9/9$ identities improved), with paired identity-bootstrap 95\% confidence intervals for Monolithic-minus-Full MAE entirely above zero (five-group diagnostics in Sec.~\ref{sec:supp_review_controls}).
 
\noindent\textbf{Reconstruction and Update Cost.}
On the formal benchmarks (Table~\ref{tab:ablation}(b)), global-local decomposition reduces active-mask mouth-closure MAE by \textbf{3.30\%} on VRoid ($32/48$ identities) and \textbf{4.31\%} on zero-shot Genshin ($20/24$). Meanwhile, with global controls $p_g$ fixed, caching global anchors accelerates local mouth, brow, and gaze updates to \textbf{78.64 FPS} versus $59.39$ FPS for monolithic decoding. Despite monolithic decoding retaining slightly better full-frame reconstruction and higher full-motion throughput (Sec.~\ref{sec:supp_ablation}), our decomposition significantly improves closure responsiveness, preserves unedited regions, and accelerates local updates.
 
\noindent\textbf{Angular Anchor Budget.}
Under matched decoder-only recovery on 64 training identities and eager-mode timing, increasing $K$ from $12$ to $40$ lowers LPIPS ($0.1903\to0.1736$) but reduces throughput ($22.88\to16.15$ FPS) and increases peak memory ($1.077\to2.484$~GiB). We select $K=12$ as a fidelity-cost compromise (Table~\ref{tab:supp_anchor_budget}; supervision ablations in Sec.~\ref{sec:supp_ablation}).

\section{Conclusion}
\label{sec:conclusion}

We presented Ref2Light for feed-forward, display-native light-field animation from a single frontal A-pose illustration and continuous head-body and facial controls.
It combines cached identity features, pose-conditioned angular anchors, correspondence-guided facial refinement, and calibrated subpixel interlacing without per-character 3D reconstruction or full-resolution $V$-view rendering.
Ref2Light improves multi-view fidelity on unseen VRoid and native-raster accuracy on zero-shot Genshin over 2D/3D baselines; matched comparisons with monolithic decoding show more accurate mouth closure and substantially lower non-target change errors.
On a consumer GPU, Ref2Light generates GPU-ready native rasters in real time with a small memory footprint and supports rapid switching between pre-aligned character inputs within a warm session.

\section*{LLM Usage Statement}
Gemini 4 Argon was used to assist with language editing, and GPT-6 Astra was used to assist with code implementation. All outputs were reviewed and verified by the authors, who take full responsibility for the contents of this paper.

{
    \small
    \bibliographystyle{ieeenat_fullname}
    \bibliography{main}
}

\appendix



\section{Additional Method Details}
\label{sec:supp_method}

This section details the architectural specifications, coordinate transformations, tracking procedures, local control heads, and optimisation objectives referenced in Sec.~\ref{sec:method}.

\subsection{Input Preprocessing and Control Routing}
\label{sec:supp_input}

\textbf{Reference Preprocessing.}
Given an input 2D character illustration $R$, the preprocessing operator $\mathcal{P}$ rectifies image orientation, composites transparency onto a solid black background, and isotropically fits the illustration into a square region vertically centred within the $H_r \times W_r$ portrait canvas $C=\mathcal{P}(R)$. Subsequently, our pipeline synthesises $K<V$ refined angular views $A^* \in [0,1]^{K \times 3 \times H_a \times W_a}$, which the interlacing operator $\mathcal{S}_\kappa$ routes to the display-native light-field raster $\widehat{Y} \in [0,1]^{3 \times H_n \times W_n}$ (Sec.~\ref{sec:formulation}).

\textbf{Control Parameterisation and Partitioning.}
In our implementation, the continuous semantic control vector $p \in \mathbb{R}^{14}$ (supplied by facial tracking or animation sliders) specifies six-dimensional head and body rotation, eye openness, mouth opening $u_o$, signed mouth shape $p_8$, three-dimensional eyebrow articulation $p_j^{brow}$ ($j \in \{1,2,3\}$ for height, distance, and curvature), and 2D gaze. Specifically, for the signed mouth shape parameter $p_8$, negative values control mouth curvature while positive values control radial expansion.

To prevent interference between coarse head/body transport and localised facial articulation, we partition $p$ between the global generator $\mathcal{G}_\theta$ and local refinement operator $\mathcal{L}_{\phi,\psi}$. Prior to standardisation, the routed global vector $p_g$ retains head/body rotation, eye openness, and non-positive mouth curvature, while neutralising mouth opening, positive radial expansion, brows, and gaze. Consequently, fine articulation along these neutralised axes is delegated to the local controllers in Sec.~\ref{sec:local_ctrl} and Sec.~\ref{sec:supp_local}.

\subsection{Conditional Generator and Feature Modulation}
\label{sec:supp_global}

\textbf{Multi-Scale Identity Encoder.}
Our global generator $\mathcal{G}_\theta$ consists of a shared encoder $E_\theta$ and conditional decoder $D_\theta$ (Sec.~\ref{sec:global_gen}). Given the portrait canvas $C$, $E_\theta$ extracts six-scale identity features $F=\{F_0,\ldots,F_5\}$ via an initial convolution, two residual blocks, and five strided downsampling stages equipped with GroupNorm-SiLU residual blocks. Since $F$ depends solely on the reference canvas $C$, it is computed once per character during onboarding, cached in memory, and fed to matching decoder scales via U-Net skip connections~\cite{ronneberger2015unet}. This way, per-frame decoding updates the character's pose and expression state while reusing its cached identity representation.

\textbf{Bounded FiLM Conditioning.}
To inject the routed global controls $p_g$ into $D_\theta$, a two-layer MLP first embeds the standardised control vector, and bounded Feature-wise Linear Modulation (FiLM)~\cite{perez2018film} then modulates intermediate decoder feature maps $X$ as:
\begin{equation}
\begin{aligned}
    z &= \operatorname{MLP}((p_g-\mu)\oslash\sigma), \\
    X' &= [1+\eta\tanh\gamma(z)]\odot X+\beta(z).
\end{aligned}
\label{eq:film}
\end{equation}
where $(\mu,\sigma)$ denote dataset mean and standard deviation statistics, $\oslash$ and $\odot$ denote element-wise division and channel-wise multiplication, $\gamma$ and $\beta$ are linear projections from latent code $z$, and $\eta$ bounds the multiplicative modulation magnitude. This conditioning operates at the decoder bottleneck and across all five upsampling stages.

\textbf{Output Head and Differentiable Composition.}
At the decoder output, spatial centre-cropping, convolution, and pixel-shuffle upsampling yield six raw channels for each of the $K$ angular views: two backward coordinate offsets $U_k$, three predicted RGB colours $D_k$, and one mixing weight $\alpha_k$. Here, scaled $\tanh$ bounds the backward offsets $U_k$, while sigmoid activations bound the direct colours $D_k$ and mixing weights $\alpha_k$ to $[0,1]$. As described in Sec.~\ref{sec:global_gen}, differentiable bilinear grid sampling~\cite{jaderberg2015spatial} queries $C$ at clipped coordinates $\operatorname{clip}(g+U_k,-1.2,1.2)$ with zero-padding for out-of-bounds locations, and the final blending between warped texture $W_k$ and predicted colour $D_k$ executes in FP32. Consequently, low mixing weights $\alpha_k$ favour direct reuse of visible reference strokes in $W_k$, whereas high mixing weights favour predicted colour $D_k$ in newly exposed regions.

\subsection{Correspondence-Based Facial Tracking}
\label{sec:supp_tracking}

To keep local facial controllers attached to the moving face across all $K$ angular views without inference-time 3D meshes, we initialise and track facial bounding boxes $\mathcal{B}$ via the backward coordinate transport field (Sec.~\ref{sec:local_ctrl}). Specifically, during character onboarding, MediaPipe Face Landmarker~\cite{mediapipe2024,lugaresi2019mediapipe} first detects initial eye, mouth, and eyebrow boxes on neutral angular views synthesised with open eyes, while manual bounding boxes provide a fallback when automatic detection fails. Using the neutral backward offsets, we then map the initial eye and mouth boxes onto $C$ to cache their central source UV quantile ranges ($20\%$ to $80\%$).

Subsequently, during per-frame driving, for each angular view $k$, we search within bounded spatial windows for pixels whose transported source coordinates $g+U_k$ fall within each facial region's cached UV quantile range. The centroid shift of the matched pixels translates each view's eye and mouth bounding boxes without resizing, while median valid shifts across views locate the shared context crop. Meanwhile, for the eyebrows, tracking additionally estimates scale variations from the matched source footprints. Since region matching operates in canonical source coordinates rather than fixed screen coordinates, the tracked boxes $\mathcal{B}$ follow changes in head and body pose in each angular view purely in 2D anchor space.

\subsection{Local Facial Controllers}
\label{sec:supp_local}

\textbf{Canonical Crop Formatting and Masking.}
Given the tracked facial boxes $\mathcal{B}$, context-padded crops are extracted from the current angular views and warped reference textures and resized to canonical crop resolutions. Specifically, the gaze, primary mouth, and brow modules each take a 9-channel input tensor concatenating the current view crop $B_{\mathrm{crop}}$ (three channels), the corresponding warped reference crop (three channels), local 2D canonical spatial coordinates $g_{loc}$ (two channels), and the 1D viewing angle coordinate (one channel). Conditioning on the viewing angle allows a single set of shared network weights to serve all $K$ angular views.

Each module then predicts a bounded 2D local displacement field $u$ and a 3-channel colour residual $r$ to update the crop via Eq.~\eqref{eq:local}. In Eq.~\eqref{eq:local}, the soft mask $M$ follows a smoothstep falloff that vanishes at crop boundaries and enforces complementary spatial weights between the eye and mouth regions to prevent overlapping write-back seams. We bilinearly resize the masked residual $\Delta$ back to the tracked box dimensions, add it to the tracked view region, and clamp the result to $[0,1]$. In our pipeline, local controllers execute sequentially across gaze, primary mouth opening/curvature, positive radial mouth expansion, and eyebrows.

\textbf{Gaze and Pose-Conditioned Mouth Closure.}
The gaze controller employs a conditional U-Net with gate $a=\min(1,|p_{12}|+|p_{13}|)\operatorname{clip}(p_6,0,1)$ in Eq.~\eqref{eq:local}, suppressing gaze updates when the eyes close.

Meanwhile, in the primary mouth branch, decoder features are conditioned on mouth opening and non-positive mouth curvature to predict 5-channel closed and open endpoint fields $Q_{close,0},Q_{open,0}$ (two displacement and three colour channels each). The index $0$ denotes endpoints before the additional pose correction. To eliminate residual mouth openings during head/body motion prior to endpoint interpolation, the zero-initialised residual head $H_\psi$ described in Sec.~\ref{sec:local_ctrl} takes mouth decoder features $F_{mouth}$, current and warped-reference mouth crops $B_{mouth},W_{mouth}$, and embeddings of the six head/body rotation controls $p_{0:6}$ via channel-wise modulation. Given the normalised opening weight $t=\operatorname{clip}(u_o/u_{max},0,1)$, the pose-corrected endpoint field $Q_{mouth}=(1-t)(Q_{close,0}+\Delta Q_p)+tQ_{open,0}$ is bounded by $\tanh$ and applied via Eq.~\eqref{eq:local}. Since $\Delta Q_p$ is added in endpoint space before nonlinear $\tanh$ bounding, head/body pose can jointly adjust both sampling displacement and colour; zero initialisation preserves the uncorrected endpoints initially, while $(1-t)$ concentrates the correction near closure and vanishes at full opening ($t=1$).

\textbf{Positive Radial Mouth Expansion.}
To articulate wide-open mouth shapes without interfering with negative curvature, the positive radial expansion head reuses the mouth decoder features, the preliminary mouth-updated RGB result, and local spatio-angular coordinates. Using two hidden convolutions and an output convolution, this radial network first predicts separate closed and open radial endpoints. These endpoints are then interpolated using the same opening weight $t$, after which the displacement and colour residuals are bounded with separate magnitude caps and scaled by $r_+=\operatorname{clip}(p_8/r_{max},0,1)$, where $p_8$ is the signed mouth shape control and $r_{max}$ is its radial normalisation ceiling. As a result, radial updates vanish identically for non-positive mouth shape ($p_8 \le 0$). Executing this head after the primary mouth update ensures that positive radial expansion acts directly on the already articulated mouth, while negative mouth curvature remains routed through the global and primary mouth branches.

\textbf{Independent Brow Axis Composition.}
The eyebrow module applies two shared hidden convolutions and an output layer to the 9-channel brow crop to predict signed 5-channel endpoint fields $Q_j^-,Q_j^+$ for each of the three brow axes $j \in \{1,2,3\}$ (height, distance/spacing, and curvature). Given normalised brow controls $b_j=\operatorname{clip}(p_j^{brow}/b_{max},-1,1)$ with normalisation ceiling $b_{max}$, the composite eyebrow field $Q_{brow}$ is formed via signed linear combination as:
\begin{equation}
\begin{aligned}
    Q_{brow} &= \sum_{j=1}^{3}\max(-b_j,0)Q_j^- \\
             &\quad+\sum_{j=1}^{3}\max(b_j,0)Q_j^+.
\end{aligned}
\label{eq:brow}
\end{equation}
where signed weighting selects the active directional endpoint per axis while summation enables simultaneous articulation of height, spacing, and curvature, and zero controls yield an identically zero field before the local update. Finally, the resulting field is bounded and written back into the angular views via Eq.~\eqref{eq:local} under a unified dual-brow soft mask to complete the refined view volume $A^*$.

\subsection{Calibrated Subpixel Interlacing}
\label{sec:supp_interlace}

As described in Sec.~\ref{sec:native_output}, an autostereoscopic panel routes RGB subpixels towards $V$ physical viewing zones via a lenticular array or parallax barrier. To synthesise the display-native raster $\widehat{Y} \in [0,1]^{3 \times H_n \times W_n}$ from the $K<V$ refined angular views $A^*$, the interlacing operator $\mathcal{S}_\kappa$ indexes each physical subpixel $(x,y,c)$ using normalised spatial coordinates $(g^x(x),g^y(y)) \in [-1,1]^2$ and a channel-dependent normalised angular coordinate $g_c^z(x,y) \in [-1,1]$. Specifically, calibrated lenticular pitch, slant, offset, and RGB subpixel position determine a discrete view index for each subpixel, which is linearly mapped across the $K$ ordered views to obtain $g_c^z(x,y)$. Since trilinear interpolation ($\operatorname{TriSample}(A_c^*,g^x(x),g^y(y),g_c^z(x,y))$) is differentiable with respect to the input view volume $A^*$, this FP32 sampling enables end-to-end native-raster supervision without rendering all $V$ full-resolution views.

\subsection{Surface Correspondence and Paired Supervision}
\label{sec:supp_pair}

\textbf{Canonical Surface Projection and Ground-Truth Offsets.}
To supervise the decomposition between coordinate transport $W_k$ and appearance synthesis $D_k$ in Sec.~\ref{sec:global_gen}, we record canonical 3D surface positions $X_s(x)$ for each visible target pixel $x$ during training using a renderer with fixed per-character mesh topology (Sec.~\ref{sec:supervision}). Projecting $X_s(x)$ into the canonical square reference image yields normalised UV coordinates $q_s(x)=\pi_s(X_s(x))=(u,v)$. Accounting for the vertical centring of the square illustration within the portrait canvas $C$ of aspect ratio $\rho=W_r/H_r$, the normalised source sampling coordinate $g_s(u,v)$ and ground-truth backward offset $U^{gt}(x)$ relative to the normalised target grid $g_t(x)$ are formulated as:
\begin{equation}
\begin{aligned}
    g_s(u,v) &= (2u-1,\;2\rho v-\rho), \\
    U^{gt}(x) &= g_s(q_s(x))-g_t(x).
\end{aligned}
\label{eq:uv}
\end{equation}

\textbf{Disjoint Supervision Masks and Paired Loss Terms.}
Based on material validity, surface identity, inverse depth consistency, and displacement bounds, we classify target pixels into reliably copyable pixels $M_f$ and confirmed non-copyable pixels $M_d$, while excluding ambiguous or unverified pixels from both masks ($M_d \neq 1-M_f$). In the paired objective $\mathcal{L}_{pair}$ (Sec.~\ref{sec:supervision}), the masked error operator $\ell_M$ computes the channel-averaged absolute ($L_1$) error summed over mask $M$ and divided by the active pixel count of $M$ clamped below at one. This way, $\ell_{M_f}(U,U^{gt})$ supervises backward coordinate offsets on reliably copyable surfaces, $\ell_{M_d}(D,I^{gt})$ supervises direct colour predictions against rendered ground-truth views $I^{gt}$ on confirmed non-copyable regions, $\mathcal{L}_{rgb}$ reconstructs the rendered views $I^{gt}$, and $\mathcal{L}_{native}$ penalises foreground-weighted absolute error on sampled native subpixels against the ground-truth panel raster $Y$.

\textbf{Reference Augmentation Variants.}
During training, we apply the paired objective across three reference variants sharing identical target outputs: a neutral reference ($\mathcal{L}_o$), a geometrically augmented reference ($\mathcal{L}_a$), and an expression-varied reference ($\mathcal{L}_e$). Specifically, in $\mathcal{L}_a$, isotropic scaling, 2D translation, and lower-body cropping are applied to $C$, synchronously transforming the ground-truth correspondence coordinates in Eq.~\eqref{eq:uv} while reassigning any source regions removed by cropping to the non-copyable mask $M_d$. Meanwhile, in $\mathcal{L}_e$, using open-mouth or combined open-radial reference illustrations alters source facial appearance relative to the neutral topology; we therefore disable neutral-source geometric correspondence labels while retaining full view-space ($\mathcal{L}_{rgb}$) and native-space ($\mathcal{L}_{native}$) image supervision.

\subsection{Multi-Reference Replay and Optimisation}
\label{sec:supp_training}

\textbf{Multi-Reference Replay Objective.}
To suppress source-expression leakage and enforce consistent control responses across diverse input expressions, the replay task pairs four references per training identity (neutral, maximum mouth-open, maximum radial expansion, and combined open-radial) with shared target control states. As defined in Sec.~\ref{sec:supervision}, given the active reference-to-native mapping $\mathcal{F}$, its frozen initial copy $\mathcal{F}_{\mathrm{init}}$, and the rigid-only control counterpart $p_{\mathrm{base}}$ (retaining head/body angles, opening the eyes, and neutralising all other facial expressions), the response loss $\mathcal{L}_\Delta$ aligns $\Delta\mathcal{F}(R_a,p)$ with the frozen response $\Delta\mathcal{F}_{\mathrm{init}}(R_0,p)$, where $\Delta\mathcal{F}(R,p)=\mathcal{F}(R,p)-\mathcal{F}(R,p_{\mathrm{base}})$ denotes the control-induced response for a fixed reference $R$. Here, $R_0$ is the original reference and $R_a$ its augmented counterpart. Here, the frozen model supplies a consistent response target across references. Consequently, the full replay objective $\mathcal{L}_{rep}$ combines $\mathcal{L}_\Delta$ with full-frame, lower-body, and facial ground-truth native reconstruction losses ($\mathcal{L}_{all},\mathcal{L}_{lower},\mathcal{L}_{faceGT}$) and appearance retention losses against $\mathcal{F}_{\mathrm{init}}$ on facial and body regions ($\mathcal{L}_{faceKeep},\mathcal{L}_{bodyKeep}$) as:
\begin{equation}
\begin{aligned}
    \mathcal{L}_{rep} &= \mathcal{L}_{all}+\lambda_l\mathcal{L}_{lower}
      +\lambda_h\mathcal{L}_{faceGT} \\
      &\quad+\lambda_a\mathcal{L}_{faceKeep}
      +\lambda_\Delta\mathcal{L}_\Delta+\lambda_b\mathcal{L}_{bodyKeep},
\end{aligned}
\label{eq:replay}
\end{equation}
where $\lambda_l,\lambda_h,\lambda_a,\lambda_\Delta,\lambda_b$ weight the respective terms. Combining the three paired supervision variants with the replay objective yields the joint global loss as:
\begin{equation}
\begin{aligned}
    \mathcal{L}_{global} &= 0.2(\mathcal{L}_o+\mathcal{L}_a+\mathcal{L}_e) \\
                       &\quad+0.4\mathcal{L}_{rep}.
\end{aligned}
\label{eq:total_mix}
\end{equation}

\textbf{Staged Training and Gradient Routing.}
Prior to final native-signal refinement, training proceeds in three stages aligned with our modular decomposition:
\begin{enumerate}
    \item We first pretrain the local facial controllers $\phi$ on grouped native-subpixel states using absolute reconstruction and differential response losses.
    \item Next, we freeze the parameters $\phi$ while keeping their forward computation graph attached to retain input gradients, and warm up the pose-conditioned mouth-closure head $\psi$ with the global parameters $\theta$ held fixed.
    \item Finally, we jointly optimise $\theta$ and $\psi$ while keeping $\phi$ frozen. To refine pose-dependent mouth closure without perturbing global coordinate transport, additional mouth-region view and native closure losses $\mathcal{L}_{mouth\text{-}extra}$ are routed exclusively to $\psi$ as:
\end{enumerate}
\begin{equation}
\begin{aligned}
    g_\theta &= \nabla_\theta\mathcal{L}_{global},\qquad g_\phi=0, \\
    g_\psi &= \nabla_\psi(
      \mathcal{L}_{global}+\mathcal{L}_{mouth\text{-}extra}).
\end{aligned}
\label{eq:grad_routing}
\end{equation}
Under this gradient routing, shared supervision in $\mathcal{L}_{global}$ jointly updates global transport/synthesis ($\theta$) and mouth closure ($\psi$) as native-space gradients flow through the frozen local controllers $\phi$, while $\mathcal{L}_{mouth\text{-}extra}$ provides targeted supervision solely to the dedicated closure head $\psi$.

\textbf{Final Native-Signal Refinement.}
Starting from the pretrained 16k-stage checkpoint, our selected main model receives 2,000 refinement updates on 418 training identities. Specifically, we freeze the identity encoder and jointly optimise the decoder, local controllers, radial mouth head, and brow head using foreground-weighted native RGB $L_1$ loss (foreground/background weights $4/0.2$). Each update samples $24{,}576$ native pixel positions across full-frame/lower-body/eye/mouth regions in the ratio $3{:}1{:}1{:}1$, using AdamW with learning rate and weight decay $10^{-5}$ and gradient clipping at $1$. Reference inputs are the canonical open-mouth illustrations without geometric, colour, or crop augmentation, and the fixed final update defines the evaluated checkpoint. Building upon the preceding geometry-supervised and response-matching stages, this final stage optimises native RGB reconstruction with all local modules, including the closure head $\psi$, trainable.

\section{Additional Experimental Details and Results}
\label{sec:supp_exp}

This section reports the complete main-model comparisons, metric definitions, runtime measurements, and targeted diagnostic analyses. The selected main weights are shared across the quality, trajectory, local-control, depth-baseline, geometry, and runtime evaluations, whereas the supervision and angular-budget recovery studies follow separate controlled training protocols specified below.

\subsection{Experimental Protocol and Identity Pairing}
\label{sec:supp_protocol}

\textbf{Asset Sources and Preparation.}
VRoid assets are obtained from the VRoid Hub collection associated with PAniC-3D~\cite{chen2023panic}.
Genshin PMX assets are obtained through the AnimeRecon download script\footnote{\url{https://github.com/ShuhongChen/animerecon-benchmark/blob/master/_scripts/download_genshin_3d.sh}} and converted to VRM while retaining textures, skeletal skinning, and supported facial morphs.
We render reference images and GT views using our camera and control protocol.
Candidate assets are screened for successful import, texture availability, supported facial controls, facial visibility, and canonical input alignment.
Our 418 training identities and the formal VRoid test identities have disjoint asset IDs; Genshin assets are used for evaluation only.

\textbf{Task Protocol and Canonical Registration.}
All compared methods take the same stored, strictly frontal $512\times512$ reference illustration (with open eyes and mouth opening $0.75$, composited on black) and are evaluated across 40 physical camera directions, $540\times960$ RGB views, and the fixed $1440\times2560$ panel calibration. Reference preparation uses deterministic input-image alignment, and formal benchmark predictions are evaluated without post-inference ground-truth (GT) registration, whereas the nine-identity control diagnostic uses fixed reference-state translation (Sec.~\ref{sec:supp_controls}). GT view renders and native buffers are shared across methods and retained unchanged during scoring. For Ref2Light, facial boxes are initialised on predicted neutral anchor views; if automatic detection fails, the evaluator reuses existing prediction-derived boxes for that reference without accessing target images or character geometry.

\textbf{Task-Specific Identity Sets.}
While the final 2,000-step refinement uses 418 training identities, the held-out discrete VRoid candidate set contains 50 unseen identities across nine states (reference, head yaw $\pm10^\circ$, body yaw $\pm6^\circ$, closed eyes, closed mouth, and horizontal gaze left/right), from which main comparisons retain the fixed 48 identities completed by every supported method. On the external 34-identity Genshin candidate set, static/head-body comparisons use 22 completed paired identities, facial comparisons use 24, and local closure comparisons against THA4 use 10 identities from the external cohort. Continuous driving experiments use a preset eight-identity subset of the 48 held-out VRoid identities across three 32-frame trajectories and 40 views per frame ($30{,}720$ GT views). Throughout all main and external evaluations, the same final checkpoint is evaluated without target-domain updates or per-character optimisation.

\subsection{Baselines and Supported Control Axes}
\label{sec:supp_baselines}

We adapt official pretrained weights to the shared display calibration and supported control dimensions:
\begin{enumerate}
    \item \textbf{THA4 + Fixed Plane}~\cite{khungurn2025talking}: monocular anime animation followed by fixed-plane view generation and calibrated interlacing, evaluated on static, head-body, and facial tasks.
    \item \textbf{LHM-1B}~\cite{qiu2025lhm}: single-image Gaussian reconstruction with head-body driving, evaluated on static and head-body tasks.
    \item \textbf{CharacterGen + Pose Adaptation}~\cite{peng2024charactergen}: mesh generation with skeletal pose adaptation, evaluated on VRoid static/head-body tasks and continuous head-body motion.
    \item \textbf{THA4 + Monocular Depth}: our 2.5D adaptation pairing THA4 with Depth Anything V2 Small~\cite{depth_anything_v2}, depth-aware viewpoint generation, and the shared interlacer across the same 48/22/24-identity task cohorts as the main table (Sec.~\ref{sec:supp_depth}).
\end{enumerate}
Baseline outputs and runtime records are evaluated under identical identity, control, GT, and timing definitions, without mixing older checkpoints, while unsupported task/domain combinations are omitted.

\subsection{Complete Metric Definitions}
\label{sec:supp_metrics}

We evaluate visual quality across both the 40 decoded camera views ($540\times 960$ RGB) and the calibrated $1440\times2560$ native panel buffer:
\begin{itemize}
    \item \textbf{View-Space Fidelity Metrics:} Measured across the 40 decoded views using full-frame \textbf{PSNR}, \textbf{Foreground PSNR (FG-PSNR)} (computed over ground-truth foreground pixels where any RGB channel exceeds $0.02$), \textbf{SSIM}~\cite{wang2004ssim} (evaluated with an $11\times11$ Gaussian window, $\sigma=1.5$), and \textbf{LPIPS}~\cite{zhang2018lpips} (computed using AlexNet v0.1).
    \item \textbf{Native-Panel Accuracy Metrics:} Evaluated directly on the $1440\times 2560$ interlaced native raster image using \textbf{Native RMSE (Nat-RMSE)} and \textbf{Native Foreground MAE (Nat-MAE$_{\text{FG}}$)}.
    \item \textbf{Localised Eye/Mouth Closure MAE:} For local facial articulation, we define ground-truth active change masks as pixels where the absolute difference between the target action and canonical reference exceeds $2/255$ in any RGB channel, morphologically dilated by 5 pixels, and compute the mean absolute error within these active organ masks.
    \item \textbf{Temporal and Angular Change MAE ($\Delta_{\text{Temp}}, \Delta_{\text{Ang}}$):} For continuous 32-frame trajectories across 40 views, we define $\Delta_{\text{Temp}}$ and $\Delta_{\text{Ang}}$ as the foreground MAE between predicted and ground-truth RGB finite differences across adjacent frames (31 frame pairs) and adjacent camera views (39 view pairs), respectively.
    \item \textbf{Aggregation Protocol and Improved Identity Ratio:} To prevent bias towards specific poses or viewing angles, each metric is first averaged across all 40 views and control states within each character, and then aggregated with equal weight across the paired identities. Continuous frame-quality metrics are scored every fourth frame and at the last frame, while transition errors use all adjacent pairs. We also report the \textit{improved identity ratio}, denoting the fraction of paired identities on which our method outperforms a given competitor.
\end{itemize}

\begin{table*}[t]
\centering
\caption{\textbf{Complete six-metric quantitative comparisons on in-domain (VRoid) and zero-shot cross-dataset (Genshin) benchmarks.} All metrics are evaluated across 40 physical viewing directions and the $1440\times 2560$ native panel raster over identical paired identities completed by all methods within each task group ($N=48$ on VRoid; $N=22$ for static/head-body and $N=24$ for facial tasks on Genshin) and aggregated with equal weight per identity. Best results are in \textbf{bold}.}
\label{tab:supp_main_full}
\scriptsize
\setlength{\tabcolsep}{1.5pt}
\renewcommand{\arraystretch}{1.08}
\begin{tabular}{ll cccccc c cccccc}
\toprule
& & \multicolumn{6}{c}{\textbf{VRoid ($N=48$ paired)}} & & \multicolumn{6}{c}{\textbf{Genshin ($N=22/24$ paired)}} \\
\cmidrule(lr){3-8} \cmidrule(lr){10-15}
\textbf{Task} & \textbf{Method} & \shortstack{\textbf{PSNR}\\$\uparrow$} & \shortstack{\textbf{FG-PSNR}\\$\uparrow$} & \shortstack{\textbf{SSIM}\\$\uparrow$} & \shortstack{\textbf{LPIPS}\\$\downarrow$} & \shortstack{\textbf{Nat-RMSE}\\$\downarrow$} & \shortstack{\textbf{Nat-MAE}$_{\text{FG}}$\\$\downarrow$} & & \shortstack{\textbf{PSNR}\\$\uparrow$} & \shortstack{\textbf{FG-PSNR}\\$\uparrow$} & \shortstack{\textbf{SSIM}\\$\uparrow$} & \shortstack{\textbf{LPIPS}\\$\downarrow$} & \shortstack{\textbf{Nat-RMSE}\\$\downarrow$} & \shortstack{\textbf{Nat-MAE}$_{\text{FG}}$\\$\downarrow$} \\
\midrule
\multirow{5}{*}{\shortstack[l]{\textbf{Static}\\\textbf{Reconst.}}} 
& CharacterGen & 11.987 & 8.678 & 0.6903 & 0.4049 & 0.2569 & 0.2562 & & -- & -- & -- & -- & -- & -- \\
& LHM-1B & 12.237 & 11.411 & 0.6847 & 0.4319 & 0.2485 & 0.1796 & & 10.687 & 8.932 & 0.5404 & 0.5285 & 0.2912 & 0.2746 \\
& THA4 + Fixed Plane & 14.558 & 10.971 & 0.7746 & 0.2402 & 0.1929 & 0.1605 & & 11.966 & 8.825 & 0.6498 & 0.3508 & 0.2534 & 0.2446 \\
 & THA4 + Monocular Depth & 14.607 & 10.570 & 0.7883 & 0.2210 & 0.1934 & 0.1626 &  & 12.483 & 8.964 & 0.6792 & \textbf{0.3081} & 0.2383 & 0.2248 \\
 & \ours{} & \textbf{18.983} & \textbf{16.915} & \textbf{0.8405} & \textbf{0.1896} & \textbf{0.1174} & \textbf{0.0723} &  & \textbf{13.833} & \textbf{11.178} & \textbf{0.6831} & 0.3464 & \textbf{0.2014} & \textbf{0.1741} \\
\midrule
\multirow{5}{*}{\shortstack[l]{\textbf{Head-Body}\\\textbf{Motion}}} 
& CharacterGen & 11.897 & 8.612 & 0.6872 & 0.4115 & 0.2594 & 0.2586 & & -- & -- & -- & -- & -- & -- \\
& LHM-1B & 12.236 & 11.401 & 0.6850 & 0.4322 & 0.2485 & 0.1799 & & 10.688 & 8.923 & 0.5409 & 0.5285 & 0.2912 & 0.2750 \\
& THA4 + Fixed Plane & 14.279 & 10.778 & 0.7667 & 0.2551 & 0.1986 & 0.1659 & & 11.805 & 8.731 & 0.6433 & 0.3673 & 0.2576 & 0.2492 \\
 & THA4 + Monocular Depth & 14.190 & 10.228 & 0.7741 & 0.2401 & 0.2013 & 0.1730 &  & 12.276 & 8.827 & 0.6687 & \textbf{0.3275} & 0.2436 & 0.2320 \\
 & \ours{} & \textbf{18.911} & \textbf{16.826} & \textbf{0.8390} & \textbf{0.1918} & \textbf{0.1183} & \textbf{0.0733} &  & \textbf{13.784} & \textbf{11.133} & \textbf{0.6817} & 0.3489 & \textbf{0.2026} & \textbf{0.1753} \\
\midrule
\multirow{3}{*}{\shortstack[l]{\textbf{Eye, Mouth}\\\textbf{\& Gaze}}} 
& THA4 + Fixed Plane & 14.576 & 10.993 & 0.7751 & 0.2404 & 0.1926 & 0.1600 & & 12.113 & 8.966 & 0.6528 & 0.3482 & 0.2497 & 0.2402 \\
 & THA4 + Monocular Depth & 14.607 & 10.570 & 0.7885 & 0.2217 & 0.1934 & 0.1625 &  & 12.580 & 9.070 & 0.6810 & \textbf{0.3081} & 0.2360 & 0.2219 \\
 & \ours{} & \textbf{18.990} & \textbf{16.920} & \textbf{0.8408} & \textbf{0.1896} & \textbf{0.1173} & \textbf{0.0722} &  & \textbf{13.924} & \textbf{11.240} & \textbf{0.6844} & 0.3475 & \textbf{0.1993} & \textbf{0.1720} \\
\bottomrule
\end{tabular}
\end{table*}

\subsection{Extended Quantitative Results}
\label{sec:supp_quant}

Table~\ref{tab:supp_main_full} reports full-frame PSNR and Nat-MAE$_{\mathrm{FG}}$ alongside the four main-text metrics. As shown, Ref2Light outperforms all competitors across all six reconstruction metrics on VRoid, and achieves the best PSNR, FG-PSNR, SSIM, and native-error metrics on Genshin (while THA4 + Monocular Depth retains a lower LPIPS). Table~\ref{tab:supp_depth_full} further reports angular-change errors for this full-cohort depth comparison, while architecture controls are analysed separately below.

\textbf{Refinement Relative to the Parent.}
On VRoid head-body motion, the final native-signal refinement improves LPIPS from $0.20935$ to $0.19181$ ($8.38\%$) and Nat-RMSE from $0.12680$ to $0.11831$ ($6.70\%$), while improving Genshin head-body LPIPS from $0.37755$ to $0.34889$ ($7.59\%$) with the identity encoder, angular budget, and calibration held fixed. Table~\ref{tab:supp_architecture_full} compares the parent and matched monolithic models alongside the final model across all six metrics and task groups.

\begin{figure*}[t]
    \centering
    \includegraphics[width=\textwidth]{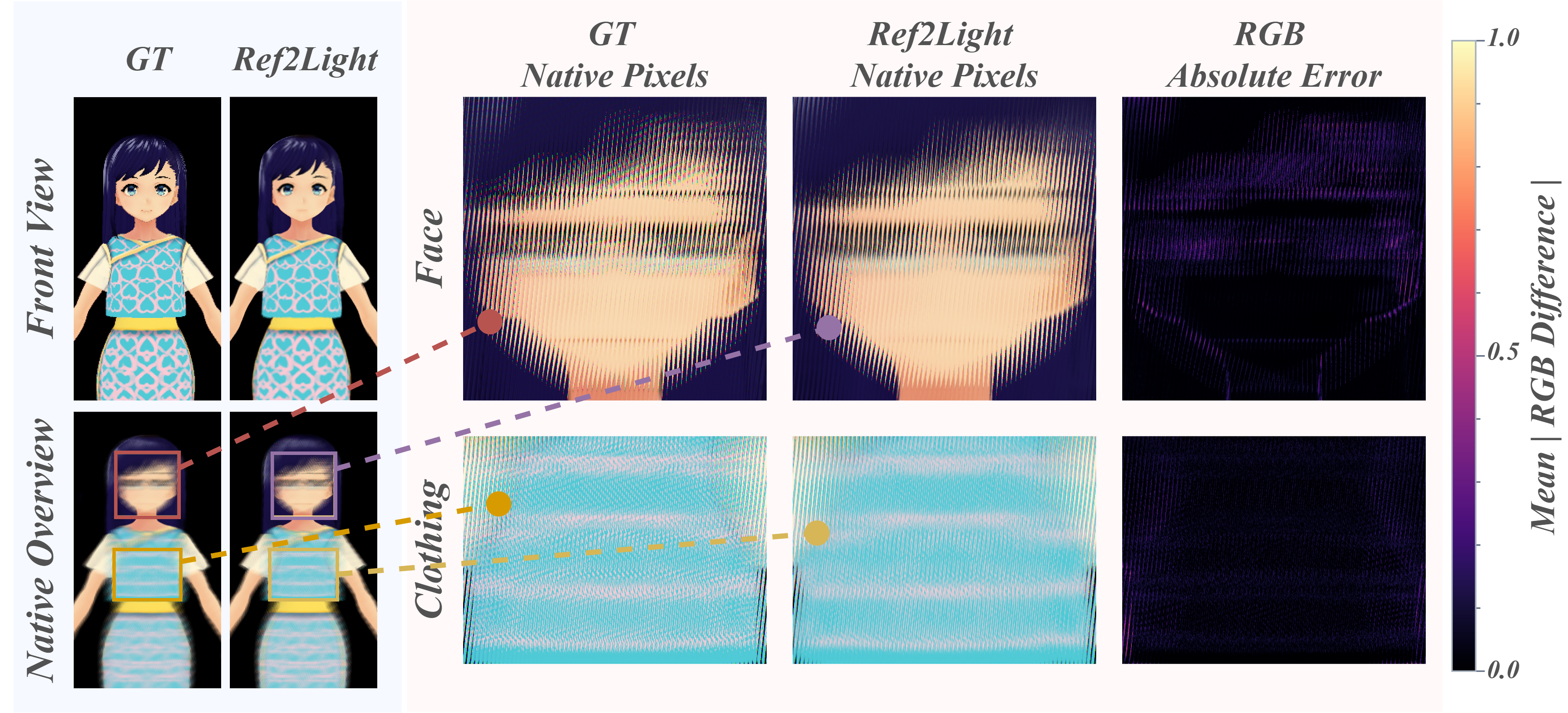}
    \caption{\textbf{Native-raster reconstruction from the selected main model.} GT and predicted native $1440\times2560$ rasters are shown with identical face/clothing pixel crops and channel-averaged absolute errors. Crops retain RGB subpixel interlacing; the shared error scale is $[0,1]$. Overview thumbnails are not used for metric computation.}
    \label{fig:supp_native_signal}
\end{figure*}

\textbf{Native-Raster Visualisation.}
Figure~\ref{fig:supp_native_signal} visually compares stored GT and final-model native RGB buffers using identical integer face/clothing pixel crops and channel-averaged absolute error maps on a shared $[0,1]$ scale (where metrics are evaluated on the full interlaced arrays while overview thumbnails and crops are used for display).

\subsection{Cross-Dataset and Local Facial Evaluation}
\label{sec:supp_cross}

On the 22 paired Genshin head-body identities, the final model achieves $0.34889$ LPIPS and $0.20256$ Nat-RMSE, reducing LPIPS and Nat-RMSE by $5.01\%$ ($15/22$ identities) and $21.37\%$ ($22/22$) over THA4, and by $33.98\%$ and $30.44\%$ ($22/22$ each) over LHM. On the 24 facial-task identities, Nat-RMSE improves by $20.18\%$ over THA4. Compared to THA4 + Monocular Depth, Genshin head-body Nat-RMSE drops by $16.84\%$ ($19/22$ identities), despite a $6.55\%$ higher LPIPS. Table~\ref{tab:supp_main_full} summarises the complete metric comparisons.

\begin{table*}[t]
\centering
\caption{\textbf{Local closure on the paired external ten-identity cohort.} MAE is measured inside GT active-change organ masks. The selected model improves both actions on all ten identities. This cohort is distinct from the full 24-identity facial-task comparison.}
\label{tab:supp_external_local}
\scriptsize
\setlength{\tabcolsep}{3.2pt}
\renewcommand{\arraystretch}{1.08}
\begin{tabular}{lccccc}
\toprule
Action & $N$ & THA4 + Fixed & Parent & Final & Reduction vs. THA4 \\
\midrule
Eye closure & 10 & 0.192838 & 0.150472 & \textbf{0.132407} & 31.34\% \\
Mouth closure & 10 & 0.253122 & 0.089181 & \textbf{0.081223} & 67.91\% \\
\bottomrule
\end{tabular}
\end{table*}

\subsection{Real-Display Capture Examples}
\label{sec:supp_real_capture}

\begin{figure*}[t]
    \centering
    \includegraphics[width=\textwidth]{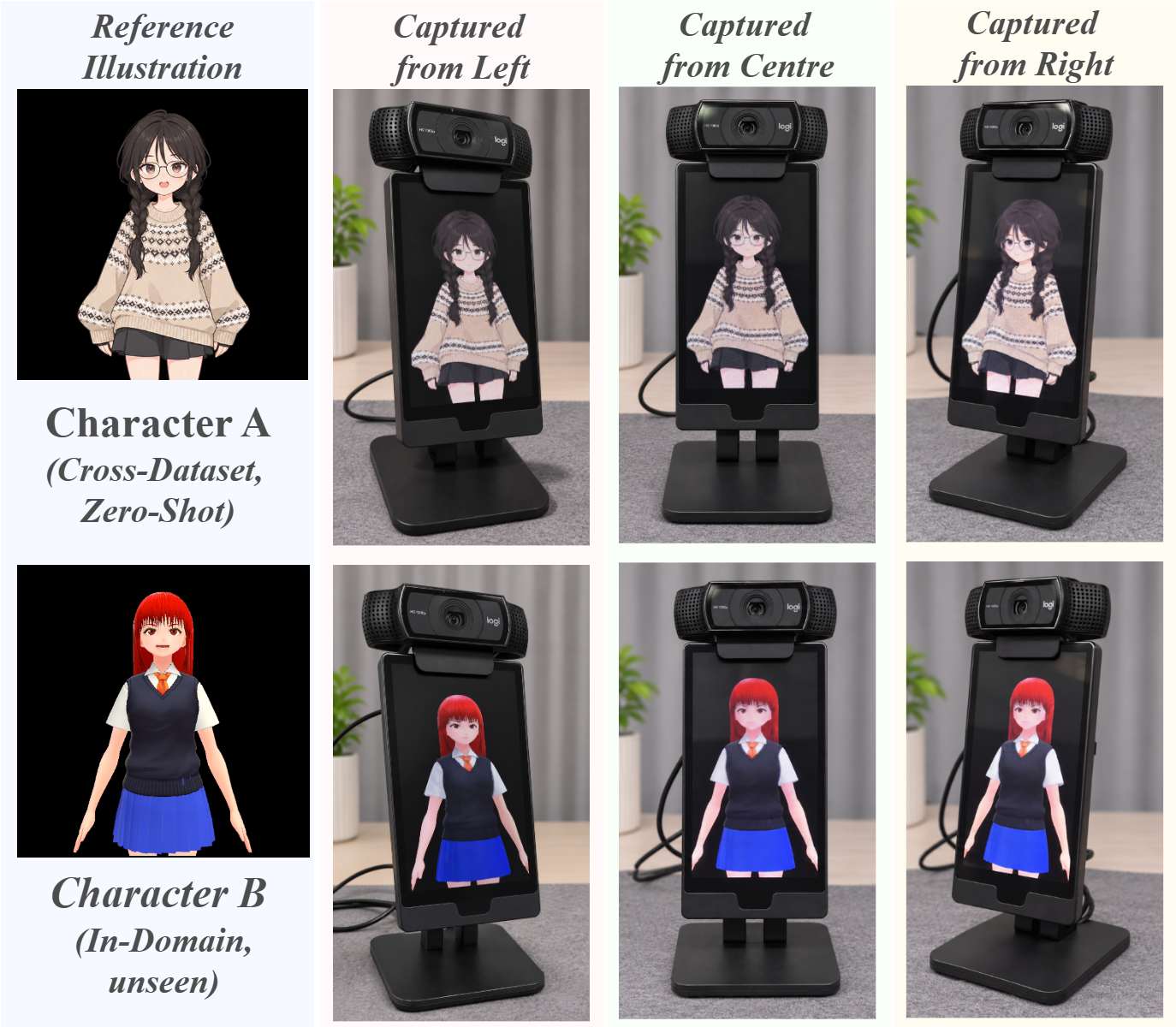}
    \caption{
        \textbf{Real-display capture examples on the physical prototype.}
        For each character, the first column shows the reference illustration, followed by the prototype viewed from the left, centre, and right.
    }
    \label{fig:supp_real_capture}
\end{figure*}

Figure~\ref{fig:supp_real_capture} complements our multi-view evaluations with photographs of character presentation on the physical light-field display prototype captured from three viewing positions, with only cropping and layout applied.

\subsection{Continuous Trajectory Protocol and Results}
\label{sec:supp_continuous}

The eight preset identities are evaluated across three 3-second (32-frame) continuous trajectories: (i) eye openness $1\to0\to1$, mouth opening $0.75\to0\to0.75$, and gaze $\pm0.8$; (ii) head yaw $\pm10^\circ$ with counter-rotating body yaw $\mp6^\circ$; and (iii) their combined motion. Frame reconstruction is scored at frames $0,4,\ldots,28,31$, while temporal and angular differences are evaluated across all 31 adjacent frame pairs and 39 adjacent viewing pairs per frame against GT finite differences (measuring transition accuracy rather than physical flicker or optical crosstalk).

\begin{table*}[t]
\centering
\caption{\textbf{Complete continuous-control results ($N=8$).} Reconstruction and GT-referenced RGB-transition errors are averaged with equal weight per identity. Lower LPIPS, Nat-RMSE, and transition MAE are better; higher PSNR/SSIM are better.}
\label{tab:continuous}
\scriptsize
\setlength{\tabcolsep}{3.2pt}
\renewcommand{\arraystretch}{1.08}
\begin{tabular}{llcccccc}
\toprule
Trajectory & Method & PSNR & SSIM & LPIPS & Nat-RMSE & $\Delta_{\mathrm{Temp}}$ & $\Delta_{\mathrm{Ang}}$ \\
\midrule
Eye, mouth, gaze & THA4 + Fixed & 14.4613 & 0.7748 & 0.241542 & 0.193971 & 0.001330 & 0.030266 \\
Eye, mouth, gaze & \ours{} & 18.7193 & 0.8327 & 0.193847 & 0.117535 & 0.004234 & 0.032287 \\
Head-body & THA4 + Fixed & 14.2649 & 0.7688 & 0.252049 & 0.197661 & 0.019865 & 0.030857 \\
Head-body & LHM-1B & 12.2637 & 0.6906 & 0.434544 & 0.244715 & 0.015876 & 0.032852 \\
Head-body & CharacterGen & 12.2179 & 0.6964 & 0.411878 & 0.247190 & 0.026538 & 0.046578 \\
Head-body & \ours{} & 18.6740 & 0.8310 & 0.195313 & 0.118127 & 0.017812 & 0.033036 \\
Combined & THA4 + Fixed & 14.2961 & 0.7698 & 0.252154 & 0.197111 & 0.019807 & 0.030340 \\
Combined & \ours{} & 18.6945 & 0.8318 & 0.195137 & 0.117890 & 0.017888 & 0.032481 \\
\bottomrule
\end{tabular}
\end{table*}

\textbf{Reconstruction and Transition Accuracy.}
Ref2Light outperforms all baselines across all four reconstruction metrics on every trajectory. For transition errors, Ref2Light achieves lower temporal-change MAE than THA4 on head-body and combined motion, whereas THA4 retains lower facial temporal error and lower angular errors across all three trajectories, and LHM yields lower head-body transition errors (while CharacterGen exhibits the highest head-body errors in both dimensions), separating per-frame reconstruction fidelity from GT-referenced RGB transition accuracy.

\begin{figure*}[t]
    \centering
    \includegraphics[width=\textwidth]{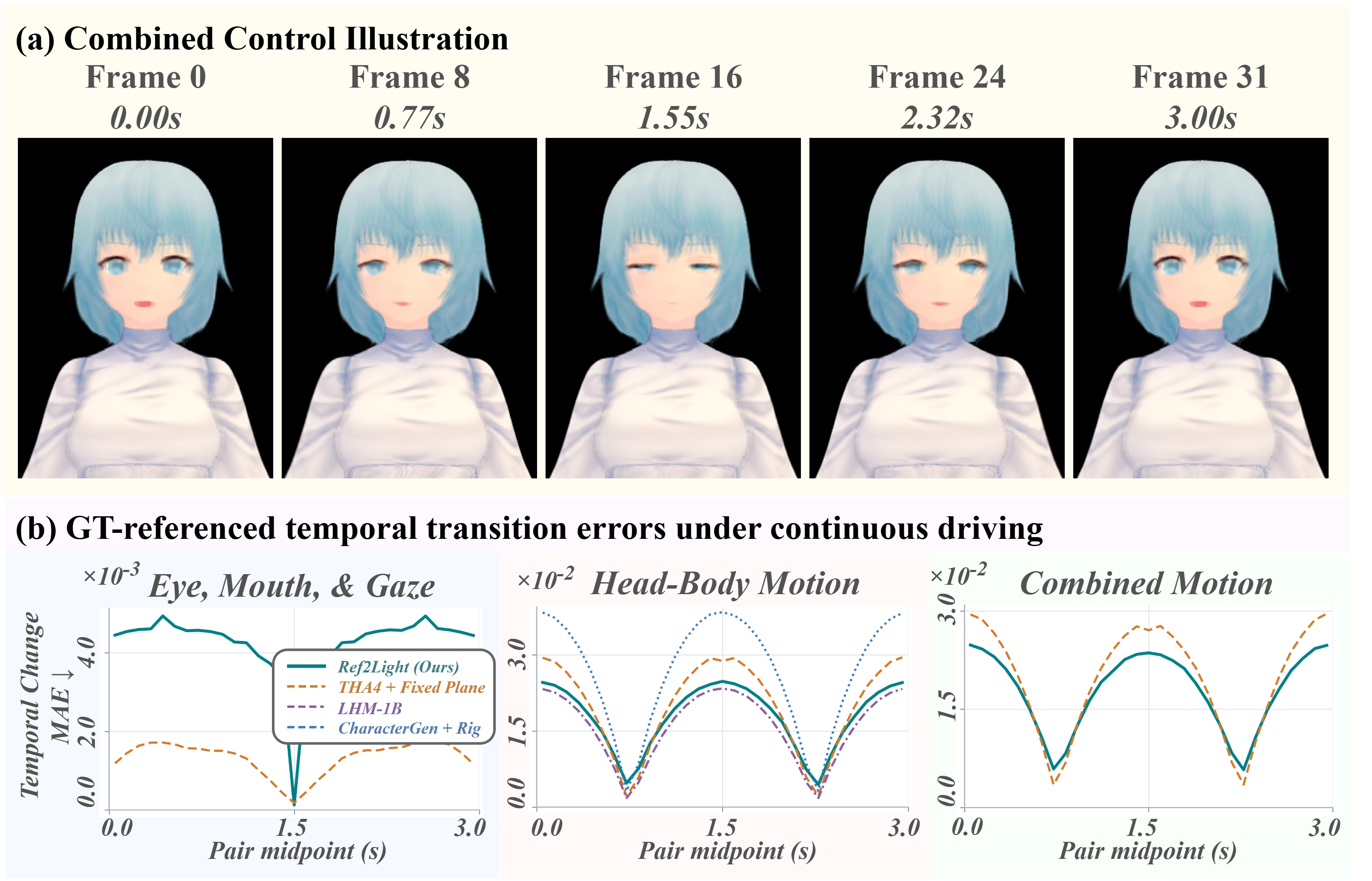}
    \caption{
    \textbf{Continuous driving illustration and GT-referenced temporal transition errors.}
    (a) Central-view outputs of Ref2Light for one training character at frames 0, 8, 16, 24, and 31 of a 32-frame combined-control sequence.
    (b) Temporal Change MAE for facial, head-body, and combined driving on eight paired test identities.
    Errors compare predicted and GT RGB differences over 31 adjacent-frame pairs, aggregated over 40 views with equal weight per identity.
    Points are plotted at frame-pair midpoints; lower values indicate closer agreement with GT changes.
}
    \label{fig:supp_continuous}
\end{figure*}

\subsection{Runtime, Onboarding, and Caching Analysis}
\label{sec:supp_runtime}

\textbf{Timing and Precision.}
Each sustained-driving condition runs eight identities, 30 warm-up frames, and 200 measured frames per identity within a single-model process. Synchronous wall-clock latency is measured from CPU dispatch to GPU-ready native RGB, with FPS defined as $1000$ divided by the equal-identity mean latency in milliseconds. CNN layers execute in BF16 where enabled, whereas coordinate sampling, RGB composition, and native interlacing execute in FP32. Our optimised path reuses identity features, caches global context when routed global inputs remain fixed, and employs CUDA Graphs, with reported allocated and reserved memory representing PyTorch peaks averaged across identities.

\begin{table*}[t]
\centering
\caption{\textbf{Same-model execution comparison.} Local controls hold every routed global input fixed, including head/body pose, eye openness, and non-positive mouth curvature. The local result does not describe full head-body/facial driving. Reserved memory is reported only where saved in the corresponding summary.}
\label{tab:supp_caching}
\scriptsize
\setlength{\tabcolsep}{3.2pt}
\renewcommand{\arraystretch}{1.08}
\begin{tabular}{lccccc}
\toprule
Condition & Path & Sync ms & FPS & Allocated GiB & Reserved GiB \\
\midrule
Eye, mouth, gaze & eager & 43.229 & 23.133 & 1.0343 & -- \\
Eye, mouth, gaze & optimised & 28.964 & 34.525 & 1.2265 & 2.5159 \\
Head-body & eager & 37.685 & 26.536 & 1.0343 & -- \\
Head-body & optimised & 24.942 & 40.093 & 1.2260 & 2.5159 \\
Combined & eager & 42.386 & 23.593 & 1.0343 & -- \\
Combined & optimised & 29.394 & 34.020 & 1.2260 & 2.5247 \\
Local controls & optimised & 12.717 & 78.635 & 1.3688 & 3.0874 \\
\bottomrule
\end{tabular}
\end{table*}

\textbf{Character Switching.}
Evaluated across eight strictly aligned references with model weights and the execution environment pre-loaded, one-time reference preparation (file reading, canvas setup, identity encoding, neutral-view localisation, and context initialisation) averages $700.396$~ms, while first-frame native synthesis adds $26.262$~ms. Consequently, their within-trial sum averages $726.659$~ms (median $661.458$~ms, 95th percentile $981.102$~ms), where GPU-resident identity encoding alone averages $9.287$~ms (excluding external image generation/frontalisation, UI display, CPU transfer, and disk export).

\textbf{Numerical Execution Check.}
Across 1,024 paired eager and optimised executions on eight identities and four driving conditions, 1,016 native uint8 outputs are identical, while the remaining eight differ at a single channel value by one intensity level (with a maximum floating-point deviation of $3.785\times10^{-5}$). Thus, execution optimisation preserves the tested outputs up to negligible numerical differences without claiming universal bit-exactness.

\textbf{Direct Native Sampling.}
Given identical pre-generated anchors across eight identities (10 warm-up and 100 timed runs each), direct calibrated sampling requires only $0.6298$~ms and $0.0412$~GiB incremental peak allocated memory, compared to $2.1568$~ms ($3.42\times$ faster) and $0.2736$~GiB when expanding anchors to 40 full-resolution views before interlacing (with a maximum floating-point difference of ${\sim}1.07\times10^{-6}$ and minor uint8 rounding differences), isolating the sampling-stage speedup.

\begin{table}[t]
\centering
\caption{\textbf{Native sampling stage.} Latency and incremental allocation exclude anchor generation and use identical anchor volumes and calibration.}
\label{tab:supp_sampling}
\scriptsize
\setlength{\tabcolsep}{3.2pt}
\renewcommand{\arraystretch}{1.08}
\begin{tabular}{lcc}
\toprule
Path & Mean ms & Incremental GiB \\
\midrule
Direct sampling & 0.6298 & 0.0412 \\
40 views then interlace & 2.1568 & 0.2736 \\
\bottomrule
\end{tabular}
\end{table}

\textbf{Matched Precision and Execution Comparison.}\label{sec:supp_review_precision}

We additionally compare THA4 FP32/BF16 autocast and Full eager/optimised execution across the same eight identities, using 30 warm-up and 200 synchronised measured head-body frames with condition order rotated per identity. All conditions reuse cached references and measure latency up to GPU-ready native RGB, using the same official teacher weights for both THA4 precision settings and the same selected weights for both Full execution paths.

\begin{table}[t]
\centering
\caption{\textbf{Matched head-body precision and execution comparison ($N=8$).} Native RMSE uses three paired GT states per identity, separately from the full quality benchmark. These new measurements are reported separately from Table~\ref{tab:supp_caching}.}
\label{tab:supp_review_precision}
\scriptsize
\setlength{\tabcolsep}{3.2pt}
\renewcommand{\arraystretch}{1.08}
\begin{tabular}{lccc}
\toprule
Condition & Sync ms$\downarrow$ & FPS$\uparrow$ & Nat-RMSE$\downarrow$ \\
\midrule
THA4 + Fixed, FP32 & 110.810 & 9.024 & 0.194550 \\
THA4 + Fixed, BF16 & 73.537 & 13.599 & 0.193925 \\
Full, eager & 32.907 & 30.389 & 0.117835 \\
Full, optimised & 23.283 & 42.951 & 0.117835 \\
\bottomrule
\end{tabular}
\end{table}

On the 24-state quality check (three paired GT states per identity), FP32 and BF16 THA4 yield mean native RMSEs of $0.194550$ and $0.193925$ (with a maximum per-state difference of $0.000832$), confirming comparable native RMSE on these states without claiming pixelwise equivalence (THA4 is timed without CUDA Graphs).

\subsection{Depth-Enhanced 2.5D Baseline}
\label{sec:supp_depth}

THA4 + Monocular Depth pairs the official THA4 animator with Depth Anything V2 Small to introduce viewpoint-dependent parallax prior to native interlacing. We evaluate the exact same task-specific cohorts as the main comparison (48 VRoid identities per task, 22 Genshin static/head-body identities, and 24 Genshin facial-task identities) using the selected main checkpoint for both reconstruction and angular-change metrics.

Specifically, each frame is first animated by THA4, from which Depth Anything predicts relative inverse depth using $518\times518$ input and ImageNet normalisation, normalised by the 5th and 95th foreground percentiles. A single affine depth-scale calibration is fitted on eight training identities and held fixed during testing without using test geometry. Three inverse-reprojection iterations then query the 40 camera matrices, and linear-RGB bilinear sampling synthesises the multi-view images before calibrated native interlacing, forming a single-layer 2.5D adaptation rather than a full 3D reconstruction model.

\begin{table*}[t]
\centering
\caption{\textbf{Complete depth-baseline comparison on the main cohorts.} All seven metrics use the same task-specific identities as Table~\ref{tab:supp_main_full}: $N=48$ on VRoid and $N=22/24$ on Genshin. $\Delta_{\mathrm{Ang}}$ is GT-referenced foreground RGB-change MAE over 39 adjacent view pairs.}
\label{tab:supp_depth_full}
\scriptsize
\setlength{\tabcolsep}{3.2pt}
\renewcommand{\arraystretch}{1.08}
\begin{tabular}{llllccccccc}
\toprule
Domain & Task & $N$ & Method & PSNR$\uparrow$ & FG-PSNR$\uparrow$ & SSIM$\uparrow$ & LPIPS$\downarrow$ & Nat-RMSE$\downarrow$ & Nat-MAE$_{\mathrm{FG}}\downarrow$ & $\Delta_{\mathrm{Ang}}\downarrow$ \\
\midrule
VRoid & static & 48 & THA4 + Depth & 14.607 & 10.570 & 0.788342 & 0.220973 & 0.193412 & 0.162588 & 0.031810 \\
VRoid & static & 48 & \ours{} & 18.983 & 16.915 & 0.840502 & 0.189598 & 0.117396 & 0.072305 & 0.026143 \\
VRoid & head-body & 48 & THA4 + Depth & 14.190 & 10.228 & 0.774135 & 0.240137 & 0.201259 & 0.172950 & 0.032729 \\
VRoid & head-body & 48 & \ours{} & 18.911 & 16.826 & 0.838988 & 0.191810 & 0.118308 & 0.073292 & 0.026361 \\
VRoid & facial & 48 & THA4 + Depth & 14.607 & 10.570 & 0.788468 & 0.221707 & 0.193412 & 0.162545 & 0.031591 \\
VRoid & facial & 48 & \ours{} & 18.990 & 16.920 & 0.840803 & 0.189626 & 0.117317 & 0.072156 & 0.025936 \\
Genshin & static & 22 & THA4 + Depth & 12.483 & 8.964 & 0.679198 & 0.308085 & 0.238292 & 0.224807 & 0.050643 \\
Genshin & static & 22 & \ours{} & 13.833 & 11.178 & 0.683090 & 0.346378 & 0.201421 & 0.174060 & 0.047811 \\
Genshin & head-body & 22 & THA4 + Depth & 12.276 & 8.827 & 0.668747 & 0.327458 & 0.243580 & 0.232006 & 0.050568 \\
Genshin & head-body & 22 & \ours{} & 13.784 & 11.133 & 0.681713 & 0.348891 & 0.202560 & 0.175299 & 0.048279 \\
Genshin & facial & 24 & THA4 + Depth & 12.580 & 9.070 & 0.680972 & 0.308130 & 0.236019 & 0.221852 & 0.049661 \\
Genshin & facial & 24 & \ours{} & 13.924 & 11.240 & 0.684392 & 0.347496 & 0.199315 & 0.171956 & 0.047101 \\
\bottomrule
\end{tabular}
\end{table*}

Ref2Light outperforms THA4 + Monocular Depth across all seven means on VRoid and six of seven on Genshin. Under head-body motion, Ref2Light reduces VRoid LPIPS by $20.12\%$, VRoid Nat-RMSE by $41.22\%$, and angular-change MAE by $19.46\%$ on VRoid and $4.53\%$ on Genshin, whereas THA4 + Depth retains lower Genshin LPIPS ($0.30808/0.32746/0.30813$ vs.\ $0.34638/0.34889/0.34750$ across static/head-body/facial tasks).

\subsection{Extended Ablations and Mechanism Analysis}
\label{sec:supp_ablation}

\textbf{Two Matched Recovery Protocols.}
Both ablation families start from the same pretrained 16k-stage parent checkpoint and run for 2,000 updates under AdamW (learning rate $10^{-5}$) with matched sample ordering. First, the selected Full model and Monolithic control use 418 identities, $24{,}576$ native pixels per update, a frozen identity encoder, and the same weighted native RGB objective (Sec.~\ref{sec:supp_training}). Full updates the decoder and all local modules; Monolithic updates only the global decoder, routes all 14 controls through it, and bypasses local refinement and correspondence tracking. Second, the supervision and angular-budget studies train on 64 identities ($12{,}288$ sampled pixels per update) using decoder-only optimisation and frozen local controllers under a combined view-RGB, flow, direct-colour, and native objective (weights $1, 1.25, 0.1, 0.5$); removing a loss term modifies only this recovery stage while sharing prior pretraining, and varying $K$ linearly interpolates the parent output head along the angular axis prior to decoder optimisation.

\begin{table*}[t]
\centering
\caption{\textbf{Supervision recovery: every measured domain/task result.} All rows use $K=12$, the 64-identity decoder-only training protocol, and paired 18/10-identity evaluation. \textit{All terms} denotes the supervised recovery control rather than the selected main checkpoint.}
\label{tab:supp_supervision_all}
\scriptsize
\setlength{\tabcolsep}{3.2pt}
\renewcommand{\arraystretch}{1.08}
\begin{tabular}{llllccccccc}
\toprule
Domain & Task & $N$ & Losses & PSNR & FG-PSNR & SSIM & LPIPS & Nat-RMSE & Nat-MAE$_{\mathrm{FG}}$ & $\Delta_{\mathrm{Ang}}$ \\
\midrule
VRoid & static & 18 & All terms & 18.841 & 15.656 & 0.834664 & 0.188246 & 0.114379 & 0.078918 & 0.030618 \\
VRoid & head-body & 18 & All terms & 18.786 & 15.620 & 0.833162 & 0.190344 & 0.115122 & 0.079712 & 0.030882 \\
VRoid & expression & 18 & All terms & 18.848 & 15.662 & 0.834871 & 0.188302 & 0.114314 & 0.078767 & 0.030389 \\
Genshin & static & 10 & All terms & 13.814 & 11.090 & 0.701600 & 0.344282 & 0.201764 & 0.177676 & 0.046867 \\
Genshin & head-body & 10 & All terms & 13.753 & 11.045 & 0.699919 & 0.346722 & 0.203197 & 0.178916 & 0.047235 \\
Genshin & expression & 10 & All terms & 13.820 & 11.098 & 0.701872 & 0.344477 & 0.201632 & 0.177366 & 0.046640 \\
VRoid & static & 18 & w/o native & 18.717 & 15.815 & 0.830577 & 0.192508 & 0.115319 & 0.078407 & 0.030797 \\
VRoid & head-body & 18 & w/o native & 18.660 & 15.775 & 0.829003 & 0.194880 & 0.116135 & 0.079271 & 0.031055 \\
VRoid & expression & 18 & w/o native & 18.725 & 15.821 & 0.830856 & 0.192452 & 0.115240 & 0.078256 & 0.030559 \\
Genshin & static & 10 & w/o native & 13.780 & 11.178 & 0.697738 & 0.349743 & 0.202111 & 0.175980 & 0.047401 \\
Genshin & head-body & 10 & w/o native & 13.720 & 11.131 & 0.696119 & 0.351977 & 0.203476 & 0.177219 & 0.047759 \\
Genshin & expression & 10 & w/o native & 13.786 & 11.187 & 0.698080 & 0.349683 & 0.201983 & 0.175630 & 0.047155 \\
VRoid & static & 18 & w/o flow/direct & 18.864 & 15.695 & 0.836227 & 0.186295 & 0.113951 & 0.078518 & 0.030630 \\
VRoid & head-body & 18 & w/o flow/direct & 18.802 & 15.647 & 0.834561 & 0.188351 & 0.114754 & 0.079426 & 0.030905 \\
VRoid & expression & 18 & w/o flow/direct & 18.869 & 15.698 & 0.836419 & 0.186285 & 0.113898 & 0.078389 & 0.030398 \\
Genshin & static & 10 & w/o flow/direct & 13.789 & 11.102 & 0.701571 & 0.342315 & 0.202371 & 0.176983 & 0.046935 \\
Genshin & head-body & 10 & w/o flow/direct & 13.731 & 11.059 & 0.700057 & 0.344600 & 0.203709 & 0.178140 & 0.047267 \\
Genshin & expression & 10 & w/o flow/direct & 13.795 & 11.109 & 0.701883 & 0.342421 & 0.202234 & 0.176669 & 0.046697 \\
\bottomrule
\end{tabular}
\end{table*}

Including native supervision reduces VRoid head-body LPIPS by $2.33\%$ ($16/18$ identities) and mean Nat-RMSE by $0.87\%$ ($11/18$), though FG-PSNR and native foreground MAE do not improve on this task and omitting flow/direct supervision slightly improves several reconstruction means over this fixed recovery window. Consequently, these fixed-budget experiments demonstrate an incremental native-supervision benefit on key perceptual and native metrics rather than a uniform gain across every loss term.

\textbf{Architecture Comparison.}
Table~\ref{tab:supp_architecture_full} compares our selected Full model, its parent checkpoint, and the matched Monolithic control: while Monolithic decoding retains marginally better means across the six full-frame reconstruction metrics, Table~\ref{tab:supp_local_architecture} highlights the distinct local-control advantage where our factorised model consistently reduces active-mask closure error, particularly on mouth closure.

\begin{table*}[t]
\centering
\caption{\textbf{Complete reconstruction comparison with the parent and architecture control.} All rows use the same main evaluation identities and GT. Only the two refined models share the additional 2,000-step training budget.}
\label{tab:supp_architecture_full}
\scriptsize
\setlength{\tabcolsep}{3.2pt}
\renewcommand{\arraystretch}{1.08}
\begin{tabular}{llllcccccc}
\toprule
Domain & Task & $N$ & Model & PSNR & FG-PSNR & SSIM & LPIPS & Nat-RMSE & Nat-MAE$_{\mathrm{FG}}$ \\
\midrule
VRoid & static & 48 & Parent & 18.376 & 16.516 & 0.826167 & 0.207569 & 0.126057 & 0.078845 \\
VRoid & static & 48 & Monolithic & 19.041 & 16.948 & 0.841594 & 0.189248 & 0.116609 & 0.071579 \\
VRoid & static & 48 & \ours{} & 18.983 & 16.915 & 0.840502 & 0.189598 & 0.117396 & 0.072305 \\
VRoid & head-body & 48 & Parent & 18.318 & 16.448 & 0.824955 & 0.209347 & 0.126799 & 0.079644 \\
VRoid & head-body & 48 & Monolithic & 18.964 & 16.861 & 0.840125 & 0.191290 & 0.117583 & 0.072573 \\
VRoid & head-body & 48 & \ours{} & 18.911 & 16.826 & 0.838988 & 0.191810 & 0.118308 & 0.073292 \\
VRoid & expression & 48 & Parent & 18.386 & 16.527 & 0.826434 & 0.207573 & 0.125921 & 0.078658 \\
VRoid & expression & 48 & Monolithic & 19.027 & 16.931 & 0.841639 & 0.189236 & 0.116762 & 0.071610 \\
VRoid & expression & 48 & \ours{} & 18.990 & 16.920 & 0.840803 & 0.189626 & 0.117317 & 0.072156 \\
Genshin & static & 22 & Parent & 13.542 & 10.892 & 0.669724 & 0.376256 & 0.207921 & 0.183927 \\
Genshin & static & 22 & Monolithic & 13.884 & 11.235 & 0.684529 & 0.345168 & 0.200279 & 0.172403 \\
Genshin & static & 22 & \ours{} & 13.833 & 11.178 & 0.683090 & 0.346378 & 0.201421 & 0.174060 \\
Genshin & head-body & 22 & Parent & 13.509 & 10.865 & 0.668735 & 0.377547 & 0.208717 & 0.184732 \\
Genshin & head-body & 22 & Monolithic & 13.833 & 11.191 & 0.683213 & 0.347067 & 0.201480 & 0.173597 \\
Genshin & head-body & 22 & \ours{} & 13.784 & 11.133 & 0.681713 & 0.348891 & 0.202560 & 0.175299 \\
Genshin & expression & 24 & Parent & 13.627 & 10.945 & 0.670772 & 0.377685 & 0.205878 & 0.182020 \\
Genshin & expression & 24 & Monolithic & 13.973 & 11.301 & 0.685772 & 0.345945 & 0.198206 & 0.170256 \\
Genshin & expression & 24 & \ours{} & 13.924 & 11.240 & 0.684392 & 0.347496 & 0.199315 & 0.171956 \\
\bottomrule
\end{tabular}
\end{table*}

\begin{table}[t]
\centering
\caption{\textbf{Active-mask eye/mouth closure under matched architecture refinement.} Lower MAE is better.}
\label{tab:supp_local_architecture}
\scriptsize
\setlength{\tabcolsep}{3.2pt}
\renewcommand{\arraystretch}{1.08}
\begin{tabular}{lccccc}
\toprule
Domain & Action & $N$ & Monolithic & \ours{} & Reduction \\
\midrule
VRoid & eyes-closed & 48 & 0.107346 & 0.107208 & 0.13\% \\
VRoid & mouth-closed & 48 & 0.044271 & 0.042808 & 3.30\% \\
Genshin & eyes-closed & 24 & 0.152561 & 0.151045 & 0.99\% \\
Genshin & mouth-closed & 24 & 0.111717 & 0.106905 & 4.31\% \\
\bottomrule
\end{tabular}
\end{table}

\begin{table}[t]
\centering
\caption{\textbf{Architecture runtime under each driving condition.} Each entry uses eight identities and the same 200-frame protocol. Only the local-controls condition holds all global inputs fixed.}
\label{tab:supp_architecture_runtime}
\scriptsize
\setlength{\tabcolsep}{3.2pt}
\renewcommand{\arraystretch}{1.08}
\begin{tabular}{lcccc}
\toprule
Condition & Model & Sync ms & FPS & Allocated GiB \\
\midrule
expression & Monolithic & 16.584 & 60.301 & 0.7510 \\
expression & \ours{} & 28.964 & 34.525 & 1.2265 \\
head-body & Monolithic & 16.840 & 59.383 & 0.7510 \\
head-body & \ours{} & 24.942 & 40.093 & 1.2260 \\
combined & Monolithic & 16.897 & 59.181 & 0.7510 \\
combined & \ours{} & 29.394 & 34.020 & 1.2260 \\
local-controls & Monolithic & 16.837 & 59.394 & 0.7510 \\
local-controls & \ours{} & 12.717 & 78.635 & 1.3688 \\
\bottomrule
\end{tabular}
\end{table}

Moreover, while the Monolithic control achieves higher full-motion throughput with lower allocated memory, holding global inputs fixed allows our factorised local updates to reduce latency from $16.837$ to $12.717$~ms ($24.47\%$), accelerating throughput from $59.39$ to $78.64$ FPS and confirming a clear local-precision and conditional-update trade-off.

\begin{figure*}[t]
    \centering
    \includegraphics[width=\textwidth]{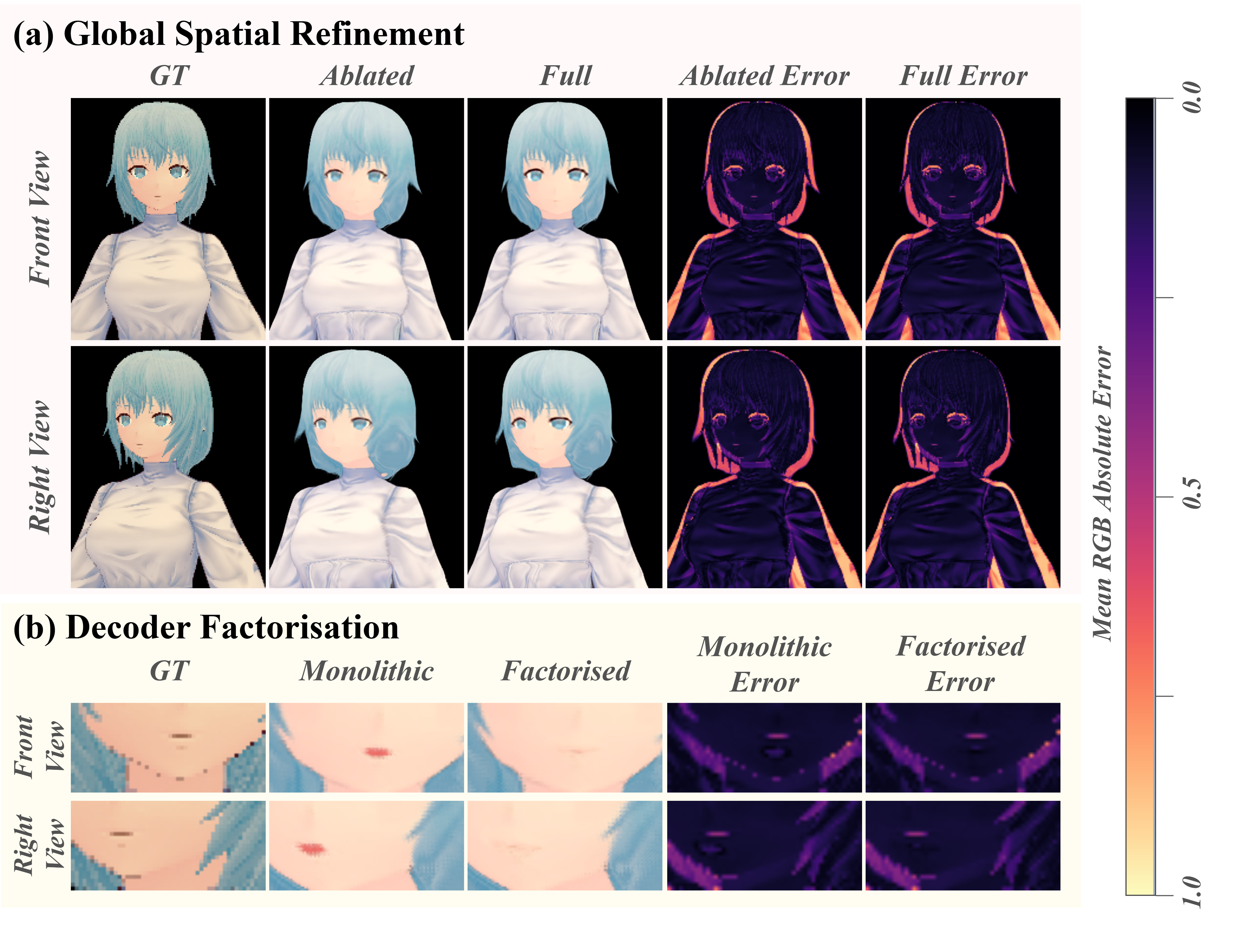}
    \caption{
        \textbf{Qualitative ablations of global spatial refinement and decoder factorisation.}
        (a) Selected frontal and right views from the global spatial-refinement ablation ($N=18$), comparing GT, the ablated variant, the full variant, and their error maps.
        (b) Mouth-region comparisons between monolithic and factorised decoding ($N=48$).
        Within each row, variants share the same character, control state, viewing position, and crop.
        Error maps show channel-averaged absolute RGB differences from GT on a shared $[0,1]$ scale.
    }
    \label{fig:supp_ablation_visual}
\end{figure*}

\noindent\textbf{Qualitative Ablation Results.}
Figure~\ref{fig:supp_ablation_visual} visually compares global spatial refinement and decoder factorisation through matched predictions and GT error maps.

\subsection{Angular Anchor Budget}
\label{sec:supp_anchor_budget}

We evaluate the $K \in \{6, 12, 20, 40\}$ models under the same 64-identity, decoder-only 2,000-step recovery protocol across 18 paired VRoid head-body identities, measuring matched eager-mode runtime across eight identities (20 warm-up and 100 timed frames per model with rotated per-identity method order and separate single-model memory profiling).

\begin{table}[t]
\centering
\caption{\textbf{Angular-budget trade-off under matched eager execution.} Changing $K$ changes the interpolated output head before the same fixed recovery budget; the models are not independently trained to convergence.}
\label{tab:supp_anchor_budget}
\scriptsize
\setlength{\tabcolsep}{3.2pt}
\renewcommand{\arraystretch}{1.08}
\begin{tabular}{lccccc}
\toprule
$K$ & LPIPS & Nat-RMSE & Sync ms & FPS & Allocated GiB \\
\midrule
6 & 0.199137 & 0.117916 & 42.988 & 23.262 & 0.7741 \\
12 & 0.190344 & 0.115122 & 43.698 & 22.884 & 1.0768 \\
20 & 0.184281 & 0.114919 & 48.690 & 20.538 & 1.4785 \\
40 & 0.173612 & 0.115981 & 61.926 & 16.148 & 2.4837 \\
\bottomrule
\end{tabular}
\end{table}

\begin{table*}[t]
\centering
\caption{\textbf{All angular-budget quality results.} All four budgets are shown for every measured domain/task group under the same decoder-only recovery protocol; quality uses paired 18/10-identity subsets.}
\label{tab:supp_anchor_quality_all}
\scriptsize
\setlength{\tabcolsep}{3.2pt}
\renewcommand{\arraystretch}{1.08}
\begin{tabular}{llllccccccc}
\toprule
Domain & Task & $N$ & $K$ & PSNR & FG-PSNR & SSIM & LPIPS & Nat-RMSE & Nat-MAE$_{\mathrm{FG}}$ & $\Delta_{\mathrm{Ang}}$ \\
\midrule
VRoid & static & 18 & 12 & 18.841 & 15.656 & 0.834664 & 0.188246 & 0.114379 & 0.078918 & 0.030618 \\
VRoid & head-body & 18 & 12 & 18.786 & 15.620 & 0.833162 & 0.190344 & 0.115122 & 0.079712 & 0.030882 \\
VRoid & expression & 18 & 12 & 18.848 & 15.662 & 0.834871 & 0.188302 & 0.114314 & 0.078767 & 0.030389 \\
Genshin & static & 10 & 12 & 13.814 & 11.090 & 0.701600 & 0.344282 & 0.201764 & 0.177676 & 0.046867 \\
Genshin & head-body & 10 & 12 & 13.753 & 11.045 & 0.699919 & 0.346722 & 0.203197 & 0.178916 & 0.047235 \\
Genshin & expression & 10 & 12 & 13.820 & 11.098 & 0.701872 & 0.344477 & 0.201632 & 0.177366 & 0.046640 \\
VRoid & static & 18 & 6 & 18.587 & 15.491 & 0.819904 & 0.197151 & 0.117078 & 0.083794 & 0.029684 \\
VRoid & head-body & 18 & 6 & 18.527 & 15.451 & 0.818264 & 0.199137 & 0.117916 & 0.084659 & 0.029917 \\
VRoid & expression & 18 & 6 & 18.597 & 15.502 & 0.820228 & 0.197138 & 0.116981 & 0.083597 & 0.029453 \\
Genshin & static & 10 & 6 & 13.825 & 11.126 & 0.694408 & 0.345536 & 0.201059 & 0.178730 & 0.042658 \\
Genshin & head-body & 10 & 6 & 13.761 & 11.078 & 0.692747 & 0.347701 & 0.202517 & 0.180021 & 0.042973 \\
Genshin & expression & 10 & 6 & 13.832 & 11.135 & 0.694773 & 0.345521 & 0.200913 & 0.178377 & 0.042429 \\
VRoid & static & 18 & 20 & 18.858 & 15.654 & 0.839738 & 0.182107 & 0.114189 & 0.077892 & 0.031137 \\
VRoid & head-body & 18 & 20 & 18.804 & 15.617 & 0.838281 & 0.184281 & 0.114919 & 0.078664 & 0.031402 \\
VRoid & expression & 18 & 20 & 18.864 & 15.657 & 0.839926 & 0.182160 & 0.114147 & 0.077772 & 0.030904 \\
Genshin & static & 10 & 20 & 13.734 & 10.985 & 0.703067 & 0.338615 & 0.203853 & 0.179287 & 0.048534 \\
Genshin & head-body & 10 & 20 & 13.675 & 10.941 & 0.701472 & 0.340801 & 0.205221 & 0.180503 & 0.048891 \\
Genshin & expression & 10 & 20 & 13.740 & 10.993 & 0.703350 & 0.338702 & 0.203716 & 0.178977 & 0.048295 \\
VRoid & static & 18 & 40 & 18.773 & 15.580 & 0.840889 & 0.171603 & 0.115172 & 0.077921 & 0.033437 \\
VRoid & head-body & 18 & 40 & 18.712 & 15.533 & 0.839182 & 0.173612 & 0.115981 & 0.078820 & 0.033798 \\
VRoid & expression & 18 & 40 & 18.778 & 15.584 & 0.841045 & 0.171639 & 0.115126 & 0.077805 & 0.033212 \\
Genshin & static & 10 & 40 & 13.650 & 10.912 & 0.701098 & 0.331188 & 0.205439 & 0.180362 & 0.053177 \\
Genshin & head-body & 10 & 40 & 13.589 & 10.863 & 0.699437 & 0.333547 & 0.206886 & 0.181696 & 0.053563 \\
Genshin & expression & 10 & 40 & 13.656 & 10.920 & 0.701348 & 0.331301 & 0.205296 & 0.180034 & 0.052949 \\
\bottomrule
\end{tabular}
\end{table*}

\begin{table}[t]
\centering
\caption{\textbf{All matched head-body eager runtime measurements from the recovery study.} Every method uses the same eight-identity timing protocol. Allocated peaks come from independent single-model processes. These rows are separate from the selected main model and its optimised runtime.}
\label{tab:supp_recovery_runtime_all}
\scriptsize
\setlength{\tabcolsep}{3.2pt}
\renewcommand{\arraystretch}{1.08}
\begin{tabular}{lccc}
\toprule
Method & Sync ms & FPS & Allocated GiB \\
\midrule
$K=12$ & 43.698 & 22.884 & 1.0768 \\
$K=6$ & 42.988 & 23.262 & 0.7741 \\
$K=20$ & 48.690 & 20.538 & 1.4785 \\
$K=40$ & 61.926 & 16.148 & 2.4837 \\
$K=12$, w/o native & 53.629 & 18.647 & 1.0768 \\
$K=12$, w/o flow/direct & 58.757 & 17.019 & 1.0768 \\
THA4 + Depth & 167.724 & 5.962 & 1.9698 \\
\bottomrule
\end{tabular}
\end{table}

While increasing $K$ consistently improves view-space LPIPS, native RMSE reaches its minimum at $K=20$ before rising slightly at $K=40$; we therefore select $K=12$ as an effective latency and memory compromise (noting that these eager-mode recovery timings are distinct from the selected main model's optimised $34$--$40$ FPS deployment speed).

\subsection{Geometry-Field Diagnostics}
\label{sec:supp_geometry}

Across 255 target views from five validation identities meeting the $2.5$-pixel input-only landmark registration threshold (with source UV labels undergoing the same affine transform and target renders unchanged), we evaluate source-canvas backward EPE on reliable, visible, and reachable regions alongside direct RGB MAE on confirmed non-copyable regions.

\begin{table}[t]
\centering
\caption{\textbf{All geometry-field diagnostic means.} The two decoder-only recovery variants form the controlled comparison. The selected main model is reported separately because its refinement protocol differs.}
\label{tab:supp_geometry}
\scriptsize
\setlength{\tabcolsep}{3.2pt}
\renewcommand{\arraystretch}{1.08}
\begin{tabular}{lcc}
\toprule
Model & Backward EPE (px) & Direct RGB MAE \\
\midrule
All-term decoder recovery & 8.1829 & 0.176920 \\
w/o flow/direct recovery & 8.0637 & 0.162064 \\
Selected main model & 6.9719 & 0.156294 \\
\bottomrule
\end{tabular}
\end{table}

In this controlled loss-removal comparison, retaining geometry losses during the 2,000-step recovery phase does not further reduce mean EPE or direct RGB error, while the selected main model's lower diagnostic errors reflect its full refinement protocol rather than an isolated causal effect of the geometry loss.

\subsection{Fine-Grained Control Diagnostics}

\label{sec:supp_controls}

\label{sec:supp_review_controls}

This independent diagnostic evaluates nine identities across 40 control states and three views, comparing the common 16k-stage Parent checkpoint with the matched 418-identity, 2,000-step refined Full and Monolithic models. For each model and view, eye geometry at the open-mouth reference state defines a single translation held fixed across all actions without action-specific registration, keeping these diagnostic scores separate from the unregistered full-frame benchmark.

Control-response errors are evaluated on the translated $540\times960$ views using a GT action mask that thresholds channel-maximum RGB change at $2/255$ with $11\times11$ dilation. Specifically, Action reconstruction MAE measures target-image error within the action mask; Action-change MAE compares predicted and GT RGB changes relative to each action's pose-matched baseline state (preserving head/body pose while holding the tested facial axis at its reference value); Outside MAE measures this change error on foreground pixels outside the action mask; Direction cosine and Amplitude ratio compare the direction and norm ratio (ideal value $1$) of the change vectors; and Zero-response MAE records the GT change magnitude under zero predicted change.

\begin{table*}[t]
\centering
\caption{\textbf{Complete control-response diagnostic ($N=9$).} All five action groups and the common parent are shown. Recon./Change/Outside MAE are lower-is-better; cosine is higher-is-better; amplitude is compared with 1. Each identity receives equal weight.}
\label{tab:supp_controls}
\scriptsize
\setlength{\tabcolsep}{3.2pt}
\renewcommand{\arraystretch}{1.08}
\begin{tabular}{llcccccc}
\toprule
Action & Model & Recon. MAE & Change MAE & Outside MAE & Cosine & Amplitude & Zero-response MAE \\
\midrule
Brow & Parent & 0.150186 & 0.046320 & 0.000098 & 0.1041 & 0.4386 & 0.027474 \\
Brow & Monolithic & 0.153695 & 0.037592 & 0.003296 & 0.0151 & 0.2281 & 0.027474 \\
Brow & \ours{} & 0.150075 & 0.038703 & 0.000085 & 0.0757 & 0.2680 & 0.027474 \\
Frontal closure & Parent & 0.035968 & 0.029489 & 0.000310 & 0.4897 & 0.6492 & 0.028163 \\
Frontal closure & Monolithic & 0.036310 & 0.029459 & 0.004701 & 0.1625 & 0.1037 & 0.028163 \\
Frontal closure & \ours{} & 0.036341 & 0.026236 & 0.000163 & 0.5232 & 0.6420 & 0.028163 \\
Mouth curve & Parent & 0.042031 & 0.016548 & 0.001673 & 0.1289 & 0.2130 & 0.015036 \\
Mouth curve & Monolithic & 0.043446 & 0.016425 & 0.003216 & -0.0032 & 0.1185 & 0.015036 \\
Mouth curve & \ours{} & 0.043229 & 0.016372 & 0.001644 & 0.0812 & 0.2111 & 0.015036 \\
Radial opening & Parent & 0.078135 & 0.041596 & 0.000163 & 0.3265 & 0.4692 & 0.040500 \\
Radial opening & Monolithic & 0.080520 & 0.042025 & 0.003464 & -0.0059 & 0.0625 & 0.040500 \\
Radial opening & \ours{} & 0.076370 & 0.042499 & 0.000155 & 0.3259 & 0.4759 & 0.040500 \\
Posed closure & Parent & 0.037121 & 0.030132 & 0.000365 & 0.4968 & 0.6199 & 0.028732 \\
Posed closure & Monolithic & 0.038514 & 0.030090 & 0.004722 & 0.1826 & 0.1071 & 0.028732 \\
Posed closure & \ours{} & 0.037759 & 0.027299 & 0.000209 & 0.5209 & 0.6239 & 0.028732 \\
\bottomrule
\end{tabular}
\end{table*}

\begin{table*}[t]
\centering
\caption{\textbf{Paired closure-control gains over Monolithic.} Gain is Monolithic minus Full MAE. The 95\% intervals resample the nine paired identities 10,000 times; wins count identities with lower Full error. These intervals describe identity variation for the fixed checkpoints.}
\label{tab:supp_control_effects}
\scriptsize
\setlength{\tabcolsep}{3.2pt}
\renewcommand{\arraystretch}{1.08}
\begin{tabular}{llcccc}
\toprule
Action & Metric & Mean gain & 95\% interval & Reduction & Wins \\
\midrule
Frontal closure & Change MAE & 0.003222 & [0.001556, 0.004970] & 10.94\% & 8/9 \\
Frontal closure & Outside MAE & 0.004537 & [0.003413, 0.006129] & 96.52\% & 9/9 \\
Posed closure & Change MAE & 0.002792 & [0.001104, 0.004642] & 9.28\% & 8/9 \\
Posed closure & Outside MAE & 0.004513 & [0.003400, 0.006085] & 95.56\% & 9/9 \\
\bottomrule
\end{tabular}
\end{table*}

Our Full model reduces frontal and posed closure-change MAE by $10.94\%$ and $9.28\%$ over Monolithic ($8/9$ identities improved in each group), while reducing non-target Outside MAE across all nine identities in all five groups by $48.87\%$--$97.42\%$. Meanwhile, brow, mouth-curve, and radial-opening target-change means remain above the zero-response reference without a consistent target-change advantage over Monolithic, separating closure responsiveness from non-target preservation.

For the complementary face-crop quality scores in Table~\ref{tab:supp_control_quality}, GT reference eye geometry defines a $256\times256$ face crop with an 88-pixel inter-ocular distance, shared across all models and actions under the fixed reference-state translation.

\begin{table*}[t]
\centering
\caption{\textbf{Face-crop reconstruction quality for every control-diagnostic group.} These scores use the three diagnostic views and fixed reference-state translation. Control-response MAEs in Table~\ref{tab:supp_controls} instead use the original-resolution views.}
\label{tab:supp_control_quality}
\scriptsize
\setlength{\tabcolsep}{3.2pt}
\renewcommand{\arraystretch}{1.08}
\begin{tabular}{llccccc}
\toprule
Group & Model & $N$ & PSNR$\uparrow$ & FG-PSNR$\uparrow$ & SSIM$\uparrow$ & LPIPS$\downarrow$ \\
\midrule
Brow & Parent & 9 & 16.4495 & 16.6598 & 0.6401 & 0.3327 \\
Brow & Monolithic & 9 & 16.5404 & 16.6938 & 0.6370 & 0.3371 \\
Brow & \ours{} & 9 & 16.3482 & 16.5975 & 0.6378 & 0.3300 \\
Frontal closure & Parent & 9 & 16.4563 & 16.6685 & 0.6406 & 0.3322 \\
Frontal closure & Monolithic & 9 & 16.5550 & 16.7133 & 0.6379 & 0.3367 \\
Frontal closure & \ours{} & 9 & 16.3607 & 16.6120 & 0.6383 & 0.3298 \\
Mouth curve & Parent & 9 & 16.4519 & 16.6656 & 0.6384 & 0.3348 \\
Mouth curve & Monolithic & 9 & 16.5341 & 16.6887 & 0.6353 & 0.3391 \\
Mouth curve & \ours{} & 9 & 16.3460 & 16.5993 & 0.6357 & 0.3327 \\
Radial opening & Parent & 9 & 16.2528 & 16.4454 & 0.6303 & 0.3360 \\
Radial opening & Monolithic & 9 & 16.2259 & 16.3330 & 0.6235 & 0.3450 \\
Radial opening & \ours{} & 9 & 16.2019 & 16.4143 & 0.6287 & 0.3332 \\
Posed closure & Parent & 9 & 16.3420 & 16.5798 & 0.6374 & 0.3363 \\
Posed closure & Monolithic & 9 & 16.3971 & 16.5659 & 0.6342 & 0.3415 \\
Posed closure & \ours{} & 9 & 16.2458 & 16.5118 & 0.6343 & 0.3352 \\
Posed open mouth & Parent & 9 & 16.2547 & 16.5129 & 0.6340 & 0.3338 \\
Posed open mouth & Monolithic & 9 & 16.3287 & 16.4745 & 0.6320 & 0.3370 \\
Posed open mouth & \ours{} & 9 & 16.2071 & 16.4688 & 0.6325 & 0.3315 \\
Reference & Parent & 9 & 16.3688 & 16.6013 & 0.6372 & 0.3301 \\
Reference & Monolithic & 9 & 16.4219 & 16.5614 & 0.6328 & 0.3360 \\
Reference & \ours{} & 9 & 16.3276 & 16.5758 & 0.6362 & 0.3265 \\
\bottomrule
\end{tabular}
\end{table*}

\subsection{Angular Structure and EPI Diagnostics}

\label{sec:supp_review_epi}

We evaluate the first three eligible identities in each domain's static/head-body paired ordering (VRoid 5558202476649691603, 2814993017888526888, and 5634565105774328162; Genshin: Kazuha, Klee, and Ningguang) under the reference and head-left states across all 40 physical views for GT, Full, Monolithic, THA4 + Depth, and THA4 + Fixed Plane, using horizontal scanlines fixed prior to scoring at $18\%$, $35\%$, and $60\%$ of image height (rows 173, 336, and 575).

Stacking each fixed scanline across all 40 views to form epipolar-plane images (EPIs), Table~\ref{tab:supp_review_epi} reports RGB MAE on GT foreground pixels at the three scanlines alongside GT-referenced angular-change MAE over the foreground union of adjacent views (using the same adjacent-view RGB finite-difference definition restricted to the fixed rows, averaged across states per identity and then equally across the three identities per domain to quantify image-space angular structure rather than geometric disparity, optical crosstalk, or viewing comfort).

\begin{table*}[t]
\centering
\caption{\textbf{Fixed-scanline and full-frame angular diagnostics.} All four predicted methods are retained on the same six identities and twelve states. Every error is lower-is-better. The last column uses all image rows on this subset, rather than the full benchmark cohort.}
\label{tab:supp_review_epi}
\scriptsize
\setlength{\tabcolsep}{3.2pt}
\renewcommand{\arraystretch}{1.08}
\begin{tabular}{llcccc}
\toprule
Domain & Method & $N$ & Scanline RGB MAE & Scanline angular MAE & Full-frame angular MAE \\
\midrule
VRoid & \ours{} & 3 & 0.058414 & 0.031340 & 0.033759 \\
VRoid & Monolithic & 3 & 0.057579 & 0.031190 & 0.033734 \\
VRoid & THA4 + Depth & 3 & 0.080613 & 0.036924 & 0.038224 \\
VRoid & THA4 + Fixed Plane & 3 & 0.095998 & 0.031270 & 0.031828 \\
Genshin & \ours{} & 3 & 0.172848 & 0.041609 & 0.047531 \\
Genshin & Monolithic & 3 & 0.169260 & 0.041174 & 0.047326 \\
Genshin & THA4 + Depth & 3 & 0.172096 & 0.038781 & 0.044135 \\
Genshin & THA4 + Fixed Plane & 3 & 0.213740 & 0.050030 & 0.054177 \\
\bottomrule
\end{tabular}
\end{table*}

Finally, the central rays extracted from the exported off-axis projection matrices (with parallel optical axes) span $39.9692^\circ$ horizontally; Table~\ref{tab:supp_anchor_angles} lists the resulting $K=12$ anchor coordinates and azimuth angles under the network-to-physical view ordering, characterising simulated camera coverage rather than the physical panel's optical viewing range.
\begin{table*}[t]
\centering
\caption{\textbf{Angular anchor placement from the exported camera matrices.} Physical camera indices range from 0 to 39; fractional coordinates interpolate the ordered camera rays. Angles are horizontal central-ray azimuths in degrees.}
\label{tab:supp_anchor_angles}
\scriptsize
\setlength{\tabcolsep}{3.2pt}
\renewcommand{\arraystretch}{1.08}
\begin{tabular}{ccc}
\toprule
Anchor index & Physical camera coordinate & Azimuth ($^\circ$) \\
\midrule
0 & 39.000 & -19.985 \\
1 & 35.455 & -16.570 \\
2 & 31.909 & -13.030 \\
3 & 28.364 & -9.386 \\
4 & 24.818 & -5.664 \\
5 & 21.273 & -1.894 \\
6 & 17.727 & 1.894 \\
7 & 14.182 & 5.664 \\
8 & 10.636 & 9.386 \\
9 & 7.091 & 13.030 \\
10 & 3.545 & 16.570 \\
11 & 0.000 & 19.985 \\
\bottomrule
\end{tabular}
\end{table*}

\end{document}